\documentclass[10pt,twocolumn,letterpaper]{article}

\usepackage[]{iccv}

\usepackage[normalem]{ulem}
\usepackage{float}
\usepackage{placeins}

\definecolor{myred}{RGB}{212,63,76}
\definecolor{mypurple}{HTML}{7A70A8}
\definecolor{mygray}{HTML}{868e96}
\definecolor{mygreen}{HTML}{548C8C}
\definecolor{myorange}{HTML}{D7795C}
\definecolor{themeblue100}{HTML}{10355C}
\definecolor{themeblue80}{HTML}{4975A4}
\definecolor{themeblue70}{HTML}{1B528D}
\definecolor{themeblue30}{HTML}{B3C7DD}
\definecolor{themegreen80}{HTML}{548C8C}
\definecolor{iccvblue}{rgb}{0.21,0.49,0.74}
\definecolor{fbblue}{rgb}{0.25,0.34,0.57}
\definecolor{metablue}{rgb}{0,0.39,0.875}
\definecolor{umdred}{rgb}{0.887,0.097,0.199}
\definecolor{wikiblue}{HTML}{3566C7}
\definecolor{urlblue}{HTML}{023394}

\newcommand{\gray}[1]{\textcolor{mygray}{#1}}
\newcommand{\noindentnewline}{\\[5pt]}

\usepackage{MnSymbol}
\usepackage{rotating}
\usepackage{tabulary}
\usepackage{bbm}
\usepackage{colortbl}

\newcolumntype{x}[1]{>{\centering\arraybackslash}p{#1pt}}
\newcolumntype{y}[1]{>{\raggedright\arraybackslash}p{#1pt}}
\newcolumntype{z}[1]{>{\raggedleft\arraybackslash}p{#1pt}}
\newlength\savewidth\newcommand\shline{\noalign{\global\savewidth\arrayrulewidth
        \global\arrayrulewidth 1pt}\hline\noalign{\global\arrayrulewidth\savewidth}}
\newcommand{\tablestyle}[2]{\setlength{\tabcolsep}{#1}\renewcommand{\arraystretch}{#2}\centering\footnotesize}

\ExplSyntaxOn
\newcommand{\mytexttt}[1]{
    \small\texttt{
        \tl_set:Nn \l_tmpa_tl { #1 }
        \tl_replace_all:Nnn \l_tmpa_tl {~}{\hspace{0.3em}}
        \tl_use:N \l_tmpa_tl
    }
}
\ExplSyntaxOff

\ExplSyntaxOn
\newcommand{\mytextttforfootnote}[1]{
    \footnotesize\texttt{
        \tl_set:Nn \l_tmpa_tl { #1 }
        \tl_replace_all:Nnn \l_tmpa_tl {~}{\hspace{0.3em}}
        \tl_use:N \l_tmpa_tl
    }
}
\ExplSyntaxOff

\newcommand{\squishlist}{
    \begin{list}{$\bullet$}
        { \setlength{\itemsep}{0pt}      \setlength{\parsep}{3pt}
            \setlength{\topsep}{3pt}       \setlength{\partopsep}{0pt}
            \setlength{\leftmargin}{1.0em} \setlength{\labelwidth}{1em}
            \setlength{\labelsep}{0.5em} } }

        \newcommand{\squishend}{
    \end{list}
}

\usepackage{tcolorbox}

\usepackage{caption}

\usepackage{marvosym}

\newcommand{\real}{\mathbb{R}}
\newcommand{\one}{\mathbf{1}}

\newcommand{\bP}{\mathbf{P}}
\newcommand{\W}{\mathbf{W}}
\newcommand{\X}{\mathbf{X}}

\newcommand{\p}{\mathbf{p}}
\newcommand{\q}{\mathbf{q}}
\newcommand{\bt}{\mathbf{t}}

\newcommand{\x}{\mathbf{x}}

\usepackage[
    pagebackref,
    breaklinks,
    colorlinks,
    citecolor=fbblue,
    linkcolor=umdred,
    urlcolor=wikiblue,
]{hyperref}

\title{
    Zero-Shot Vision Encoder Grafting via LLM Surrogates
}

\author{
    Kaiyu Yue
    \quad
    Vasu Singla
    \quad
    Menglin Jia$^\dagger$
    \quad
    John Kirchenbauer
    \noindentnewline
    Rifaa Qadri
    \quad
    Zikui Cai
    \quad
    Abhinav Bhatele
    \quad
    Furong Huang
    \quad
    Tom Goldstein
    \noindentnewline
    University of Maryland
    \quad
    $^\dagger$Meta
    \noindentnewline
    {{\mytexttt{\href{https://github.com/facebookresearch/zero}{\textcolor{wikiblue}{https://github.com/facebookresearch/zero}}}}}
}

\begin{document}
\maketitle
%
%
%
\begin{figure}[H]  
    \centering
    \vspace{.2em}
    \includegraphics[width=1.0\linewidth]{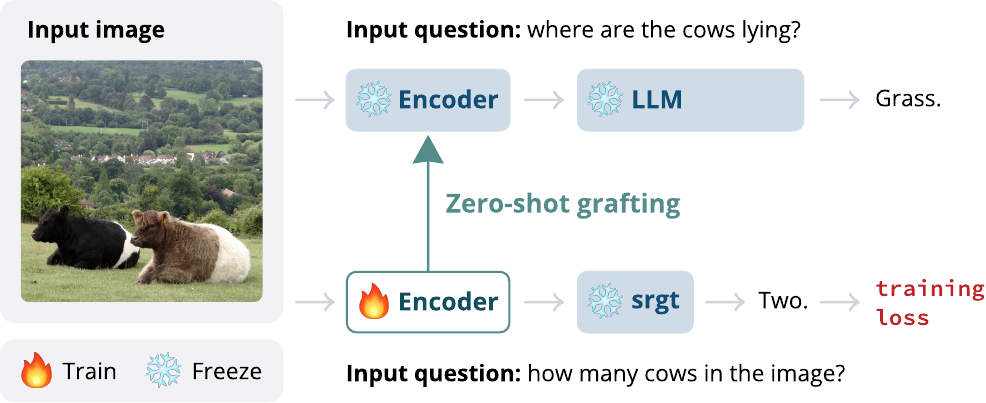}
    \vspace{-1.39em}
    \caption{
        \textbf{Zero-shot vision encoder grafting} via a small language surrogate (\textbf{srgt}) model to trigger the target LLM to perform visual understanding task without any additional training.
    }
    \label{fig:intro}
\end{figure}
%
%
\begin{abstract}
    Vision language models (VLMs) typically pair a modestly sized vision encoder with a large language model (LLM), e.g., Llama-70B, making the decoder the primary computational burden during training.
    To reduce costs, a potential promising strategy is to first train the vision encoder using a small language model before transferring it to the large one.
    We construct small ``surrogate models'' that share the same embedding space and representation language as the large target LLM by directly inheriting its shallow layers.
    Vision encoders trained on the surrogate can then be directly transferred to the larger model, a process we call zero-shot grafting\footnote{
        We define \textit{zero-shot grafting} as plugging a vision encoder trained on a surrogate model directly into its target LLM without additional training.
        In contrast, \textit{transferring} involves further fine-tuning after grafting.
    } -- when plugged directly into the full-size target LLM, the grafted pair surpasses the encoder-surrogate pair and, on some benchmarks, even performs on par with full decoder training with the target LLM.
    Furthermore, our surrogate training approach reduces overall VLM training costs by $\sim$45\% when using Llama-70B as the decoder.
\end{abstract}
%
%
%
\begin{figure}[tb] 
    \centering
    \includegraphics[width=1.\linewidth]{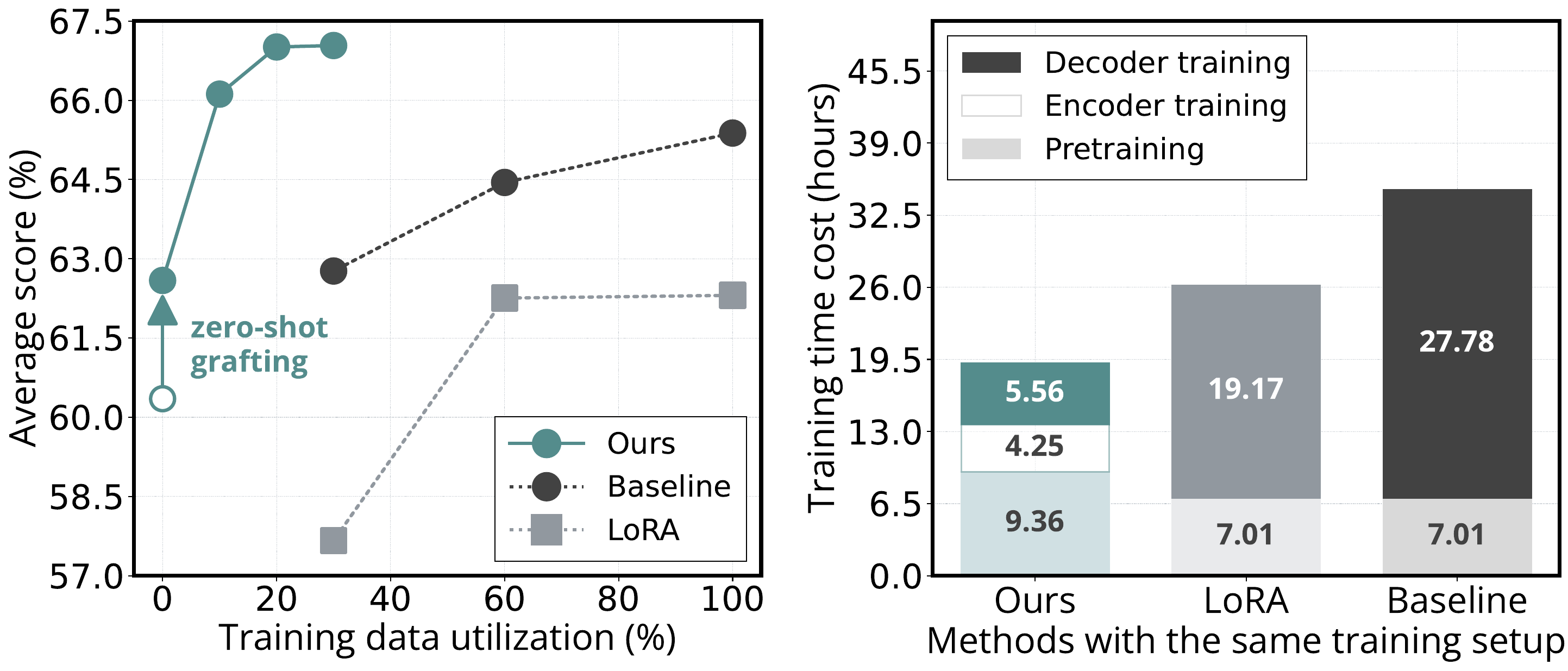}
    \vspace{-1.5em}
    \caption{
        \textbf{Reducing full decoder training cost} with our surrogate-trained encoder for Llama-70B in VLMs.
        Hollow \textcolor{themegreen80}{\CircPipe} indicates the average score of the surrogate-trained encoder on the left.
    }
    \label{fig:cost}
\end{figure}
%
%
\noindentnewline
Most modern auto-regressive VLMs are built by extracting visual features from images using an encoder like CLIP \cite{radford2021learning} or SigLIP \cite{zhai2023sigmoid,tschannen2025siglip}, and placing these features into the context window of an LLM.
The image features must be aligned with the representation space of the LLM, and this is achieved by training the entire pipeline end-to-end.
The cost of such training is often severely dominated by the language model.
For example, plugging CLIP (approx 400M parameters) into Llama-70B \cite{dubey2024llama} results in a pipeline where vision encoder training occupies almost none of the required memory and computation.
\noindentnewline
In this paper, we explore methods of performing encoder alignment using relatively small lightweight language models, and transferring the resulting features to a large language model.
We train small {\em surrogate} language models with the same representation space as a larger target LLM.
After training the vision encoder on this small surrogate model, we can then transfer it to the larger model, either directly (grafting) or with fine-tuning.
\noindentnewline
A major focus of our work is on understanding how to construct small surrogate models that accurately mock larger target LLMs.
Our method of creating such small models stems from analyzing the internal prediction dynamics of LLMs, particularly how predictions evolve across layers.
This analysis reveals two distinct phases in the prediction trajectory, separated by a clear transition point.
We construct our small models by preserving the layers that participate in the early feature extraction phase of inference, and condensing all other layers.
Since the small model inherits its shallow parameters from the target LLM, it shares the same embedding space as the original larger model and can effectively stand in as its surrogate.
Our surrogate model has two major advantages:
\noindentnewline
\textbf{Zero-shot grafting capability}.
Vision features trained on a smaller and less resource-intensive surrogate can be directly used by the larger target LLM without any fine-tuning, as depicted in Figure \ref{fig:intro}.
This {\em zero-shot grafting} demonstrates these surrogate-trained encoders effectively trigger visual understanding in target LLMs.
\noindentnewline
\textbf{Fast-converging VLM training}.
The encoders trained on surrogate models can be further fine-tuned with the full-size target LLM.
Since they are already aligned with the LLM's embedding space, they achieve high performance with comparatively little full-scale training.
Our experiments show a $\sim$45\% cost reduction for full decoder training with Llama-70B, as shown in Figure \ref{fig:cost}, highlighting the efficiency of our surrogate-trained encoders.
\begin{center}
    $\vcenter{\hbox{\rule{0.117\textwidth}{0.5pt}}}$
    \quad
    Table of Main Contents
    \quad
    $\vcenter{\hbox{\rule{0.117\textwidth}{0.5pt}}}$
\end{center}
\begin{itemize}
    \item Section~\ref{sec:method}: We detail the method of constructing our surrogate model, providing analysis that demonstrates how we discovered, developed, and validated our approach through experimental ablations.
    \item Section~\ref{sec:generalizing}: We show our surrogate models for giant LLMs like Llama-70B, producing encoders with a strong zero-shot grafting ability, which can also accelerate the full decoder training of giant language models for VLMs.
\end{itemize}
\section{Building Surrogate Models}
\label{sec:method}
In this section, we present our approach for building small surrogate models for target LLMs.
First, we analyze the LLM's hidden features to identify the critical transition point between shallow and deep information processing layers.
Next, we observe that the second/deep phase of inference contributes very little to encoder transferability, and observe that image features transfer well between models when they share their early/shallow processing layers.
Finally, we validate these findings and propose to construct surrogate models by preserving the early-phase layers while replacing late-phase layers with a translator.
%
\begin{figure*}[t]
    \centering
    \includegraphics[width=1.\linewidth]{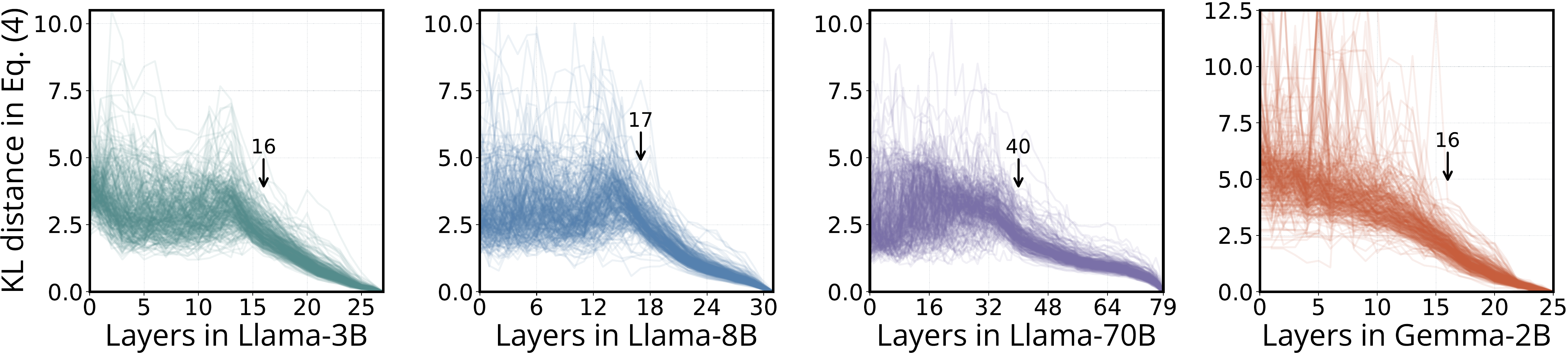}
    \vspace{-1.5em}
    \caption{
        \textbf{The trajectory of prediction} across different layers of Llama-3B, 8B, and 70B, and Gemma-2B from a different model family.
        The arrow marks the transition point where the trajectories of 300 random samples converge.
    }
    \label{fig:kl-curves}
    \vspace{-1.em}
\end{figure*}
%
\subsection{Analyzing the Prediction Trajectory}
\label{sec:prediction trajectory}
For a target LLM and input array\footnote{
    Bold capital letters denote a matrix $\X$, and bold lower-case letters a column vector $\x$.
    $\X[i, j]$ refers to the element at row $i$ and column $j$ in matrix $\X$.
    All non-bold letters represent scalars.
} of $N$ text token IDs $\bt \in \mathbb{Z}^{N}$, we trace the evolution of features over a forward pass of the model.
By propagating these tokens through all $L$ transformer layers, we obtain intermediate hidden states $\X^\ell \in \real^{N \times D}$ from each layer, where $\ell \in [0, L-1]$ denotes the layer index and $D$ is the hidden dimension.
The final hidden states $\X^{L-1}$ are passed through a normalization layer and the final linear layer $\W \in \real^{V \times D}$ to produce the logits, where $V$ is the vocabulary size.
The probability distribution for the predicted next token can be computed for all positions:
\begin{align}
    \bP = \text{softmax} \left( \text{norm}(\X^{L-1})\W^\top \right) \in \real^{N \times V}.
\end{align}
The probability for the next output token at each individual position is
\begin{align}
    \p = \bP[:-1, \bt[1:]] \in \real^{N-1},
\end{align}
where $\bP[:-1, \bt[1:]]$ shifts $\bt$ one position forward and indexes by $\bP$ up to the second-to-last position, aligning each token's probability with its following token in the sequence.
\noindentnewline
For each layer’s hidden states $\X^\ell$, we compute the intermediate probability distribution $\q^\ell$ following the same procedure:
\begin{align}
    \q^\ell = \text{softmax} \left( \text{norm}(\X^\ell)\W^\top \right)[:-1, \bt[1:]].
\end{align}
To capture the trajectory of evolving predictions, we calculate the KL divergence between the normalized layer-wise distribution $\q^\ell$ and the final distribution $\p$:
\begin{align}
    D_{KL}(\q^\ell \ || \ \p) = \one^\top (\q^\ell \log \frac{\q^\ell}{\p}),
    \label{eq:kl}
\end{align}
where $\one \in \real^{N-1}$ is a vector of ones, $\log$ is applied element-wise.
Eq.~\eqref{eq:kl} quantifies the deviation of each layer’s prediction from the final model output, offering insight into how much each layer's distribution shifts along the prediction trajectory.
This measure enables a deeper understanding of each layer's role in shaping the model's eventual output distribution.
\noindentnewline
In Figure~\ref{fig:kl-curves}, we plot Eq.~\eqref{eq:kl} across different layers of the Llama-3B, 8B, and 70B\footnote{
    Unless stated otherwise, each model mentioned refers to its latest instruct version.
    For example, Llama-3B indicates Llama-3.2 3B, Llama-70B represents Llama-3.1 70B, and Gemma-2B denotes Gemma-2 2B.
} models by feeding\footnote{
    One concern about this teacher-forced manner is ablated in Sec. \ref{sec:ableation teacher-forced}.
} 300 random samples from GenQA \cite{chen2024genqa}.
To demonstrate the same curve pattern in a different model family, Gemma-2B is also included.
Each model displays a distinct \emph{phase transition} where the curves abruptly coalesce and then monotonically converge to the final distribution.
For example, in Llama-8B, this point appears to occur around layer $17$ whereas for Llama-70B it is closer to layer $40$.
We speculate that this point marks a transition in the type of position-wise information processing occurring in the model, where the internal states shift from \textit{early phase} before the transition point to the \textit{late phase} after it.
The layers in the early phase process information from individual token embeddings and combine simple representations together to form higher order concepts, then layers in the late phase converge towards a specific next-token prediction.
%
\begin{figure}[h]
    \centering
    \includegraphics[width=1\linewidth]{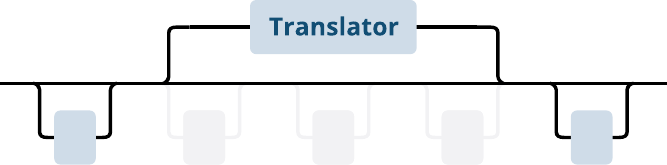}
    \vspace{-1.5em}
    \caption{
        \textbf{Replacing layers with a translator}.
        Despite the relative size in the illustration, our translator is simply an identical transformer layer inherited from the target LLM.
        The translator bypasses many network layers, and is initialized from the shallowest original layer that it replaced.
    }
    \label{fig:translator}
    \vspace{-.5em}
\end{figure}
%
%
\begin{figure*}[t]
    \centering
    \includegraphics[width=1\linewidth]{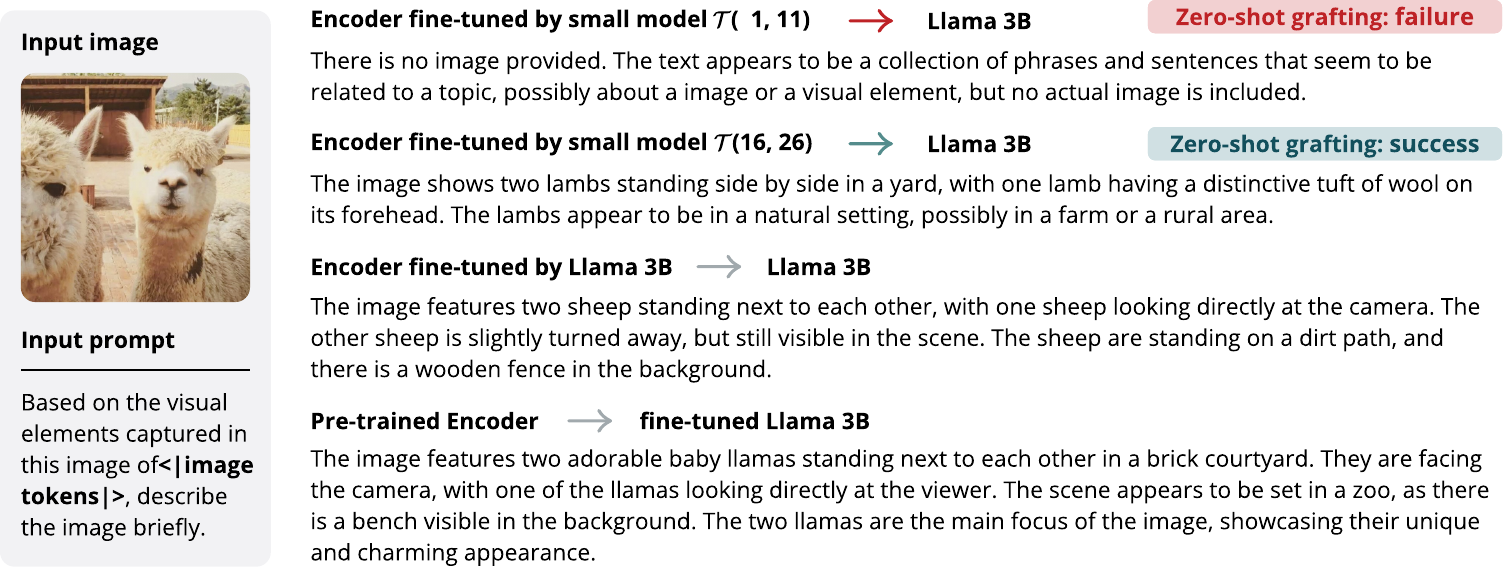}
    \vspace{-1.5em}
    \caption{
        \textbf{Qualitative results on zero-shot grafting capability} of encoders trained with small models for Llama-3B.
        For comparison, we also include responses from the encoder trained with Llama-3B and the fine-tuned Llama-3B.
        The encoder trained on $\mathcal{T}(16, 26)$ achieves strong zero-shot transfer to Llama-3B.
        Response is sampled with greedy decoding.
        A \textrightarrow \ B denotes plugging A into B.
    }
    \label{fig:zero-shot-grafting-llama-3b}
    \vspace{-.5em}
\end{figure*}
%
%
\subsection{Studying the Transition Phases}
To test our hypothesis on the transition point, we experiment with Llama-3B\footnote{
    Initial experiments with Gemma-2B showed similar results.  In later sections we adapt the method to a 70B model.
} by replacing consecutive layers of each phase with a single transformer layer called a \textit{translator} (terminology adopted from \cite{belrose2023eliciting}), as depicted in Figure~\ref{fig:translator}.
From Llama-3B's 28 layers, we preserve the first ($\ell = 0$) and last ($\ell = 27$) layers while replacing two groups of eleven layers each with a translator $\mathcal{T}$: layers from $\ell = 1$ to $11$ indicated as $\mathcal{T}(1, 11)$ for the early phase before Llama-3B's transition point ($\ell = 16$ in Figure~\ref{fig:kl-curves}) and $\mathcal{T}(16, 26)$ for the late phase after it.
Each is a 2B small model.
\noindentnewline
Next, we examine the two transition phases by evaluating vision encoders trained on models $\mathcal{T}(1, 11)$ and $\mathcal{T}(16, 26)$.
To understand their differences and how they affect encoder transferability to the target LLM, we construct two LLaVA-like VLMs using these small models as decoders.
We employ a two-stage training approach to conduct the initial experiments:

\noindent
1) First, we \textit{simultaneously} pre-train a vision adapter (a two-layer MLP) and the translator on 1M instructions\footnote{
    This differs from the typical pre-training of vision adapters, which use captions rather than instructions.
}, combining LLaVA-1.5-665K \cite{liu2024improved} vision-language instructions and random GenQA \cite{chen2024genqa} 500K text instructions, for one epoch.

\noindent
2) Then, we fine-tune the encoder ({\mytexttt{ViT-L/14@336px}}) and vision adapter with the frozen decoders on the LLaVA-1.5-665K instructions for one epoch.
%
%
\begin{table}[ht]
    \vspace{-.5em}
    \tablestyle{1.18pt}{1.05}
    \begin{tabular}{x{56}|z{18}z{18}z{18}z{18}z{18}z{18}z{18}z{18}}

        small model \ \ \               & \rotatebox{90}{mmlu}                   & \rotatebox{90}{hellaswag}  & \rotatebox{90}{arc$_\text{easy}$}
                                        & \rotatebox{90}{arc$_\text{challenge}$} & \rotatebox{90}{winogrande} & \rotatebox{90}{piqa}
                                        & \rotatebox{90}{boolq}                  & \rotatebox{90}{openbookqa}                                     \\
        \shline
        Llama-3B  \ \ \                 & 60.7                                   & 73.0                       & 71.1
                                        & 52.7                                   & 70.6                       & 77.1
                                        & 78.9                                   & 39.2                                                           \\
        \hline
        $\mathcal{T}(\ \ 1, 11)$ \ \ \  & 26.6                                   & 42.5                       & 50.3
                                        & 27.7                                   & 53.5                       & 66.6
                                        & 57.5                                   & 32.4                                                           \\
        $\mathcal{T}(16, 26)$ \ \ \     & 58.9                                   & 57.2                       & 54.8
                                        & 38.5                                   & 64.3                       & 67.6
                                        & 78.2                                   & 32.6                                                           \\
        \hline
        $\mathcal{T}(16, 26)^\ast$ \    & 56.9                                   & 57.3                       & 57.5
                                        & 40.7                                   & 64.3                       & 70.1
                                        & 79.9                                   & 35.2                                                           \\
    \end{tabular}
    \vspace{-1.em}
    \caption{
        \textbf{Accuracy (\%) of small models} for Llama-3B on text benchmarks.
        $^\ast$ is a control experiment added later in the study.
    }
    \label{tab:surrogate_3b_llm_benchmarks}
    \vspace{-1.em}
\end{table}
%
%
\noindentnewline
\textbf{Evaluating decoders}.
After fine-tuning translators in the first stage, we evaluate models $\mathcal{T}(1, 11)$ and $\mathcal{T}(16, 26)$ on text benchmarks\footnote{
    To ease the benchmarking, we evaluate our instruct models on the same benchmarks as the non-instruct models, i.e., base models, and report accuracy produced by log-likelihood.
} (Table~\ref{tab:surrogate_3b_llm_benchmarks}).
The first row is the baseline performance of Llama-3B.
The second and third rows show the performance of the decoders with early- and late-phase layers replaced, respectively.
A significant performance drop occurs when replacing early-phase layers, underscoring their critical role in understanding and generation.
\noindentnewline
\textbf{Evaluating encoders}.
During the second stage, encoders are fine-tuned with small models $\mathcal{T}(1, 11)$ and $\mathcal{T}(16, 26)$.
We also train an encoder with the full-size Llama-3B as our baseline, listed in the first row of Table~\ref{tab:surrogate_3b_vlm_benchmarks}.
For each model, $\mathcal{T}(1, 11)$ and $\mathcal{T}(16, 26)$, we report two results:
a) performance with their respective encoders, and
b) performance with these encoders \textit{zero-shot grafted} to Llama-3B.
For case b), since Llama-3B is never trained on vision-language instructions, it cannot consistently follow special instructions in benchmarks like MME \cite{fu2023mme} and POPE \cite{li2023evaluating} that expect ``yes'' or ``no'' answers by prompting with ``single word or phrase''.
For these benchmarks, we prompt the model with binary prompts, directing it to answer with ``yes'' or ``no'' to ensure measurable responses.
%
%
\begin{table}[h]
    \vspace{.5em}
    \tablestyle{1.18pt}{1.05}
    \begin{tabular}{z{76}|z{20}z{20}z{20}z{20}z{20}z{24}|x{12}}
        encoder fine-tuned on \ \ \ \           & \rotatebox{90}{MME$^\text{perc}_\text{binary}$} & \rotatebox{90}{POPE$_\text{binary}$} & \rotatebox{90}{SEED-Bench}
                                                & \rotatebox{90}{MMVet}                           & \rotatebox{90}{LLaVA-Wild}           & \rotatebox{90}{MMB$^\text{en}$} \ \ \
                                                & \rotatebox{90}{performance}                                                                                                    \\
        \shline
        Llama-3B \ \ \ \                        & 1028                                            & 81.7                                 & 54.2
                                                & 24.1                                            & 42.9                                 & 41.8 \ \ \
                                                &                                                                                                                                \\
        \hline
        model $\mathcal{T}(\ \ 1, 11)$ \ \ \ \  & 599                                             & 63.2                                 & 25.8
                                                & 14.3                                            & 37.2                                 & 0.6  \ \ \                            \\
        zero-shot grafting \ \ \ \              & 540                                             & 2.7                                  & 25.3
                                                & 6.9                                             & 26.3                                 & 9.2 \ \ \
                                                & $\downarrow$                                                                                                                   \\
        \hline
        model $\mathcal{T}(16, 26)$ \ \ \ \     & 923                                             & 70.4                                 & 53.2
                                                & 20.6                                            & 42.7                                 & 45.4 \ \ \                            \\
        zero-shot grafting \ \ \ \              & 1022                                            & 80.1                                 & 53.1
                                                & 23.1                                            & 56.6                                 & 47.4 \ \ \
                                                & $\uparrow$                                                                                                                     \\
        \hline
        model $\mathcal{T}(16, 26)^\ast$ \ \    & 1162                                            & 84.4                                 & 59.8
                                                & 25.0                                            & 48.1                                 & 50.3 \ \ \                            \\
        zero-shot grafting \ \ \ \              & 714                                             & 22.3                                 & 40.7
                                                & 11.2                                            & 34.2                                 & 30.1 \ \ \
                                                & $\downarrow$                                                                                                                   \\
    \end{tabular}
    \vspace{-1.em}
    \caption{
        \textbf{Accuracy (\%) of encoders fine-tuned by small models} for Llama-3B on VLM benchmarks.
        $^\ast$ indicates a control experiment added later in the study.
    }
    \label{tab:surrogate_3b_vlm_benchmarks}
    \vspace{-1.em}
\end{table}
%
%
\noindentnewline
Table~\ref{tab:surrogate_3b_vlm_benchmarks} clearly shows that the encoder trained with early-phase layers \emph{preserved} model $\mathcal{T}(16, 26)$ outperforms the one with early-phase layers \emph{discarded} model $\mathcal{T}(1, 11)$.
Remarkably, performance improves further when the encoder fine-tuned on $\mathcal{T}(16, 26)$ is zero-shot grafted to Llama-3B, as shown in the third row block.
This improvement highlights that the encoder trained with $\mathcal{T}(16, 26)$ can produce image features that are interpretable by Llama-3B.
\noindentnewline
In Figure~\ref{fig:zero-shot-grafting-llama-3b}, we present qualitative results showcasing the zero-shot grafting cabability of the encoders trained via $\mathcal{T}(1, 11)$ and $\mathcal{T}(16, 26)$.
The responses enhance the above results that replacing the early-phase layers causes the encoder to fail in generating image features that are directly interpretable by the full-size Llama-3B.
%
%
\\[7pt]
\textbf{Are early layers the most critical for encoder transfer?}
The shallow phase of inference plays a crucial role in transferring a pre-trained encoder to the full-size target LLM.
To concretely verify this observation, we conduct a control run based on $\mathcal{T}(16, 26)$, in which we unfreeze every other layer before the translator and train them alongside it during the first stage.
This control experiment is designed to disrupt the original early-phase parameters, allowing us to completely assess their impact on encoder transferability.
We denote this modified model as $\mathcal{T}(16, 26)^\ast$.
\noindentnewline
First, back in Table~\ref{tab:surrogate_3b_llm_benchmarks}, the last row indicates that fine-tuning additional layers alongside the translator leads to better performance on text benchmarks.
\textit{However}, in Table~\ref{tab:surrogate_3b_vlm_benchmarks}, when evaluating the encoder trained on $\mathcal{T}(16, 26)^\ast$, we actually observe a huge loss of zero-shot grafting ability.
This suggests that modifying early-phase parameters in $\mathcal{T}(16, 26)$ enhance performance on both text and VLM benchmarks when evaluated through itself, but fails to preserve the encoder's zero-shot grafting capability as the embedding space of $\mathcal{T}(16, 26)^\ast$ drifts away from the target model.
%
%
%
\begin{figure*}[t]
    \centering
    \includegraphics[width=1\linewidth]{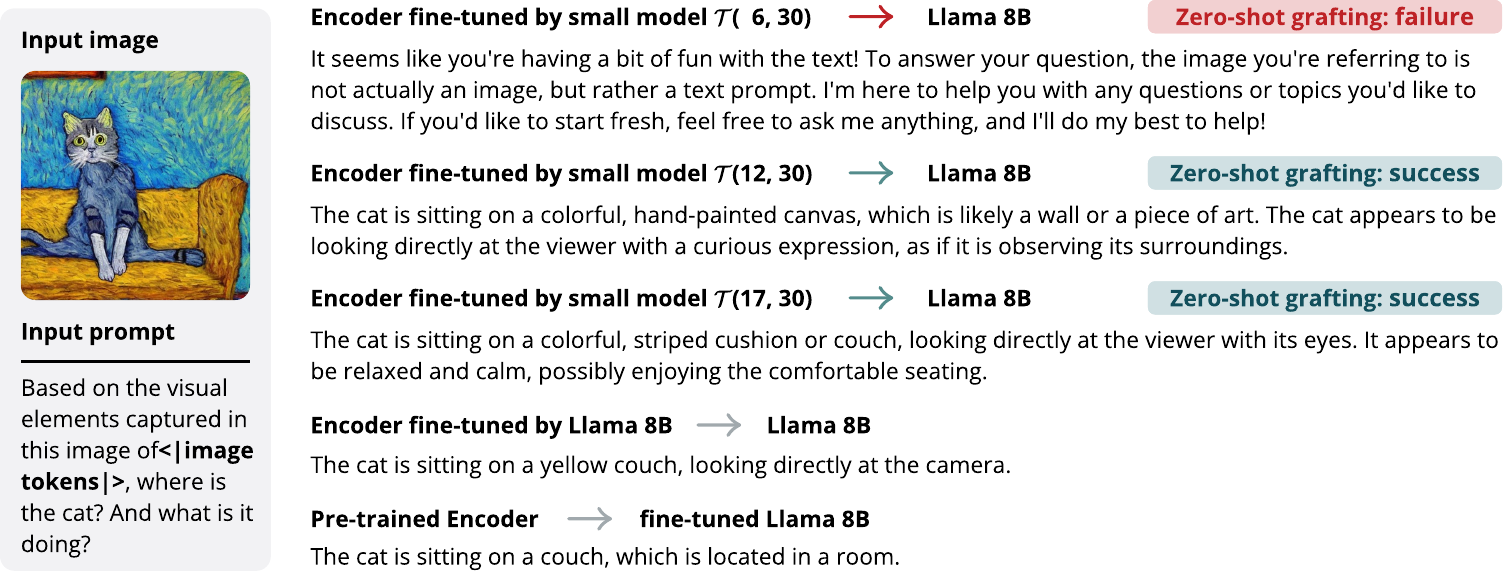}
    \vspace{-1.5em}
    \caption{
        \textbf{Qualitative results on zero-shot grafting capability} of encoders trained with surrogate models for Llama-8B.
        For comparison, we also include responses from the encoder trained with Llama-8B and the fine-tuned Llama-8B.
        More early-phase layers preserved lead to stronger zero-shot grafting capability.
        Responses are sampled with greedy decoding.
        A \textrightarrow \ B denotes plugging A into B.
    }
    \label{fig:zero-shot-grafting-llama-8b}
    \vspace{-.5em}
\end{figure*}
%
\\[7pt]
\textbf{How many early-phase layers should be preserved?}
If retaining the original early-phase parameters is necessary, the next question is how many layers to preserve for effective zero-shot grafting.
In other words, we seek to confirm the transition point in Figure~\ref{fig:kl-curves} as the optimal starting point for layer removal and translator insertion.
\noindentnewline
To ensure generalization, we conduct this ablation with Llama-8B, where the transition point is around layer $\ell = 17$.
To validate the transition point, we create three small models by progressively reducing the replaced layers before the transition point: $\mathcal{T}(6, 30)$, $\mathcal{T}(12, 30)$, and $\mathcal{T}(17, 30)$.
As shown in Table~\ref{tab:surrogate_8b_llm_benchmarks}, the performance of these three models on text benchmarks corroborates our findings from Llama-3B, demonstrating that early layers are indeed important.
Keeping more early layers leads to better performance, with the best achieved by the model $\mathcal{T}(17, 30)$.
%
%
\begin{table}[t]
    \vspace{.5em}
    \tablestyle{1.18pt}{1.05}
    \begin{tabular}{x{56}|z{18}z{18}z{18}z{18}z{18}z{18}z{18}z{18}}
        small model \ \ \               & \rotatebox{90}{mmlu}                   & \rotatebox{90}{hellaswag}  & \rotatebox{90}{arc$_\text{easy}$}
                                        & \rotatebox{90}{arc$_\text{challenge}$} & \rotatebox{90}{winogrande} & \rotatebox{90}{piqa}
                                        & \rotatebox{90}{boolq}                  & \rotatebox{90}{openbookqa}                                     \\
        \shline
        Llama-8B  \ \ \                 & 68.4                                   & 80.5                       & 79.8
                                        & 61.8                                   & 77.3                       & 81.5
                                        & 85.4                                   & 44.8                                                           \\
        \hline
        $\mathcal{T}(\ \ 6, 30)$ \ \ \  & 25.5                                   & 31.8                       & 36.8
                                        & 24.7                                   & 50.9                       & 58.8
                                        & 61.3                                   & 25.8                                                           \\
        $\mathcal{T}(12, 30)$ \ \ \     & 25.4                                   & 42.9                       & 40.5
                                        & 29.0                                   & 59.6                       & 62.6
                                        & 69.4                                   & 29.6                                                           \\
        $\mathcal{T}(17, 30)$ \ \ \     & 66.8                                   & 61.2                       & 59.3
                                        & 44.8                                   & 70.9                       & 71.0
                                        & 69.3                                   & 34.2                                                           \\
    \end{tabular}
    \vspace{-1.em}
    \caption{
        \textbf{Accuracy (\%) of small models} for Llama-8B on text benchmarks.
    }
    \label{tab:surrogate_8b_llm_benchmarks}
    \vspace{-1.em}
\end{table}
%
%
\begin{table}[h]
    \tablestyle{1.18pt}{1.05}
    \begin{tabular}{z{74}|z{20}z{20}z{20}z{20}z{20}z{24}|x{12}}
        encoder fine-tuned by  \ \ \          & \rotatebox{90}{MME$_\text{binary}^\text{perc}$} & \rotatebox{90}{POPE$_\text{binary}$} & \rotatebox{90}{SEED-Bench}
                                              & \rotatebox{90}{MMVet}                           & \rotatebox{90}{LLaVA-Wild}           & \rotatebox{90}{MMB$^\text{en}$} \ \ \
                                              & \rotatebox{90}{performance}                                                                                                    \\
        \shline
        Llama-8B \ \ \                        & 1165                                            & 84.7                                 & 57.5
                                              & 23.2                                            & 47.6                                 & 44.9 \ \ \
                                              &                                                                                                                                \\
        \hline
        model $\mathcal{T}(\ \ 6, 30)$ \ \ \  & 583                                             & 73.3                                 & 25.8
                                              & 8.9                                             & 22.4                                 & 8.6  \ \ \                            \\
        zero-shot grafting \ \ \              & 767                                             & 20.6                                 & 30.9
                                              & 13.4                                            & 28.1                                 & - \ \ \
                                              & $\downarrow$                                                                                                                   \\
        \hline
        model $\mathcal{T}(12, 30)$ \ \ \     & 983                                             & 77.9                                 & 26.9
                                              & 13.6                                            & 29.1                                 & 0.43 \ \ \                            \\
        zero-shot grafting \ \ \              & 1022                                            & 81.7                                 & 50.7
                                              & 20.5                                            & 47.9                                 & 45.4 \ \ \
                                              & $\uparrow$                                                                                                                     \\
        \hline
        model $\mathcal{T}(17, 30)$ \ \ \     & 1041                                            & 81.3                                 & 55.4
                                              & 20.9                                            & 42.0                                 & 49.7 \ \ \                            \\
        zero-shot grafting \ \ \              & 1044                                            & 83.4                                 & 56.1
                                              & 25.2                                            & 56.8                                 & 53.5 \ \ \
                                              & $\uparrow$                                                                                                                     \\
    \end{tabular}
    \vspace{-1.em}
    \caption{
        \textbf{Accuracy (\%) of encoders fine-tuned by small models} for Llama-8B on VLM benchmarks.
    }
    \label{tab:surrogate_8b_vlm_benchmarks}
    \vspace{-1.em}
\end{table}
%
\noindentnewline
We evaluate the zero-shot grafting capability of the encoders trained with these three models to Llama-8B in Table~\ref{tab:surrogate_8b_vlm_benchmarks}.
Performance improves with more preserved early layers, showing particularly strong results when retaining all early-phase layers before the transition point in $\mathcal{T}(17, 30)$.
%
%
\begin{figure*}[t]
    \centering
    \includegraphics[width=1\linewidth]{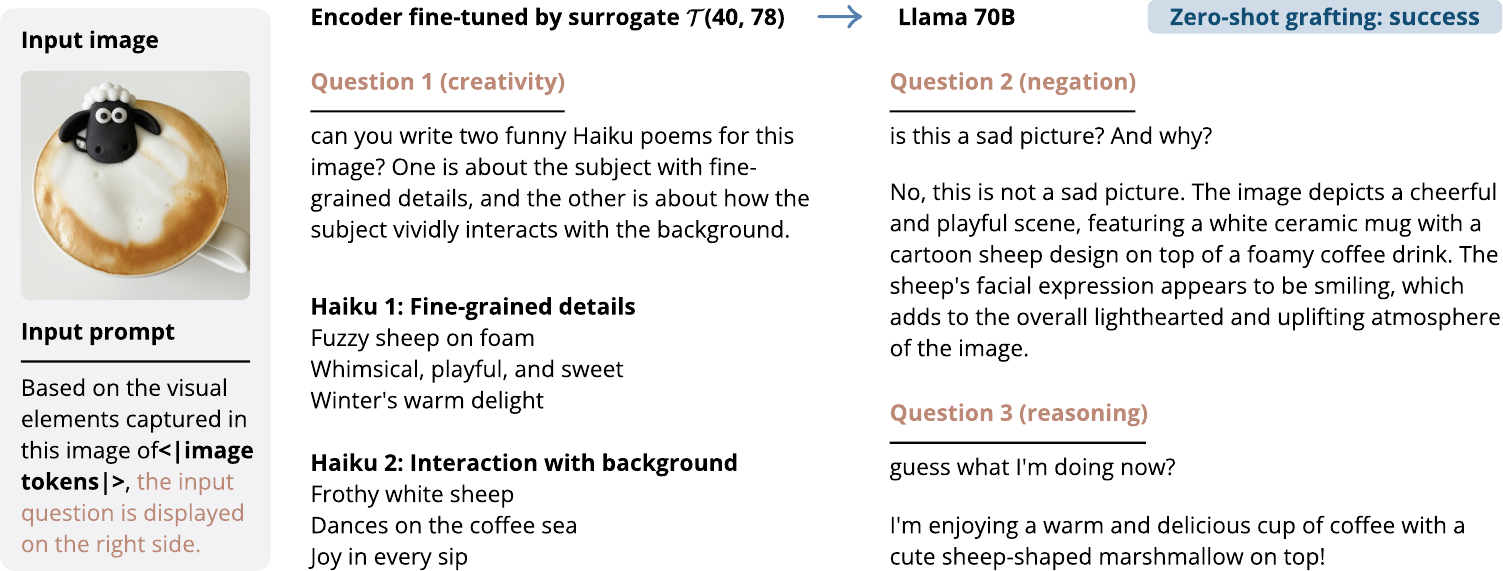}
    \vspace{-1.5em}
    \caption{
        \textbf{Qualitative results on strong zero-shot grafting ability} of surrogate-trained encoder for Llama-70B, which produces fine-grained image features to trigger Llama-70B to perform complex visual understanding tasks.
        Response is sampled with greedy decoding.
    }
    \label{fig:zero-shot-grafting-llama-70b}
    \vspace{-.5em}
\end{figure*}
%
%
\noindentnewline
Figure~\ref{fig:zero-shot-grafting-llama-8b} depicts a qualitative example demonstrating the zero-shot grafting capability of three trained encoders.
As expected, the encoder trained by the model $\mathcal{T}(6, 30)$ fails to generate readable image features for Llama-8B as most of the early-phase layers are removed.
The encoder trained by $\mathcal{T}(12, 30)$ performs better, but its image features lack fine-grained detail (e.g., no couch in response), which explains the zero-shot performance gap in Table~\ref{tab:surrogate_8b_vlm_benchmarks}.
The encoder trained by $\mathcal{T}(17, 30)$ generates more detailed and accurate image features, achieving the best zero-shot grafting response among the three, which covers the cat's color, expression, eye direction, position, the presence of a couch, and even the atmosphere.
\begin{tcolorbox}[
        fonttitle=\bfseries,
        coltitle=themeblue100!100,
        colback=themeblue30!30,
        colframe=themeblue70!0,
        width=1.0\linewidth,
        left=3pt,
        right=3pt,
        top=3pt,
        bottom=3pt,
    ]
    In summary, our entire analysis of the prediction trajectory reveals:
    \begin{itemize}
        \item The early phase\footnote{
                  We believe the early phase of LLMs has potential beyond building surrogate models, enabling more creative applications.
              } plays a pivotal role in the encoder's transferability to the target LLM.

        \item Retaining the original parameters of the early phase is critical for maintaining the encoder's zero-shot grafting capability.

        \item The transition point in Figure \ref{fig:kl-curves} is a good starting point for late-phase removal and translator insertion.
    \end{itemize}
\end{tcolorbox}
\noindent
Based on these three key findings, now we define the model $\mathcal{T}(16, 26)$ as our \textbf{\textit{surrogate model}} for Llama-3B, the model $\mathcal{T}(17, 30)$ as our surrogate for Llama-8B, by inheriting the early-phase layers and replacing the late-phase layers with a translator, which can be fine-tuned with a small set of text instructions, e.g., 500K for one epoch.
\section{Generalizing to Giant Models}
\label{sec:generalizing}
Having validated our approach at relatively small scales, we now expand our methodology to construct surrogate models for giant LLMs -- Llama-70B.
In this section, our experiments demonstrate two key advantages of our surrogate approach:
First, surrogates can bring a strong zero-shot grafting ability to encoders, enabling them to trigger target LLMs to perform visual understanding tasks without additional training.
Second, training target LLM decoders on surrogate-trained encoders significantly reduces cost by providing a warm start for fine-tuning.
\subsection{A Surrogate for Llama-70B}
We analyze the prediction trajectory for Llama-70B in Figure~\ref{fig:kl-curves} to identify the transition point that marks the end of token processing, which occurs around layer $\ell = 40$.
Then we keep the first ($\ell = 0$) and last layer ($\ell = 79$), insert a translator at $\ell = 40$, and remove the late phase from $\ell = 41$ to $\ell = 78$, to build a 37B surrogate $\mathcal{T}(40, 78)$.
Text benchmark results of this surrogate are shown in Table~\ref{tab:surrogate_70b_llm_benchmarks}, and VLM benchmark results in Table~\ref{tab:surrogate_70b_vlm_benchmarks}.
Table~\ref{tab:surrogate_70b_vlm_benchmarks} shows the performance of the encoder trained using surrogate $\mathcal{T}(40, 78)$ on VLM benchmarks, highlighting a significant improvement through zero-shot grafting.
These experiments show that our approach can be scaled up to giant models, holding the same principles of early phase preservation.
\subsubsection{Results: Zero-shot Grafting}
%
\begin{figure}[h]
    \vspace{.4em}
    \centering
    \begin{minipage}[t]{1.\linewidth}
        \includegraphics[width=1.\linewidth]{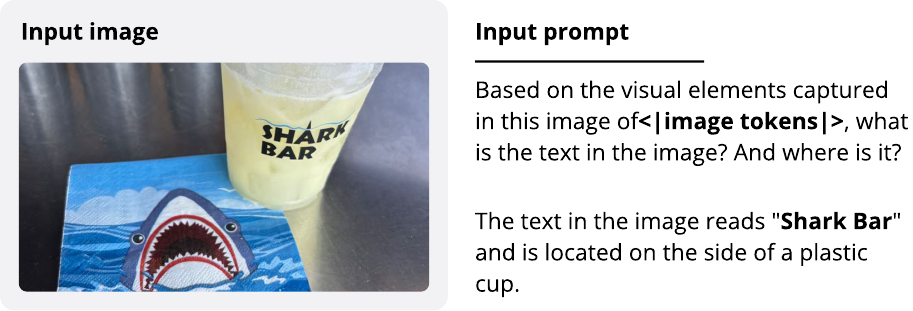}
        \vspace{-1.5em}
        \caption{
            \textbf{Qualitative OCR results on strong zero-shot grafting ability} of surrogate-trained encoder for Llama-70B.
            The input image size is 336$^2$.
        }
        \label{fig:zero-shot-grafting-llama-70b-ocr}
    \end{minipage}
    \begin{minipage}[t]{1.\linewidth}
        \captionsetup{type=table}
        \tablestyle{1.18pt}{1.05}
        \begin{tabular}{z{50}|z{18}z{18}z{18}z{18}z{18}z{18}z{18}z{18}}
            small model \ \ \    & \rotatebox{90}{mmlu}                   & \rotatebox{90}{hellaswag}  & \rotatebox{90}{arc$_\text{easy}$}
                                 & \rotatebox{90}{arc$_\text{challenge}$} & \rotatebox{90}{winogrande} & \rotatebox{90}{piqa}
                                 & \rotatebox{90}{boolq}                  & \rotatebox{90}{openbookqa}                                            \\
            \shline
            Llama-70B \ \ \      & 82.6                                   & 86.9                       & 83.4
                                 & 71.2                                   & 85.4                       & 83.7                              & 89.1
                                 & 47.6                                                                                                           \\
            \hline
            surrogate-37B \ \ \  & 80.8                                   & 70.4                       & 67.3
                                 & 56.6                                   & 77.9                       & 73.9                              & 86.9
                                 & 37.8                                                                                                           \\
        \end{tabular}
        \vspace{-1.em}
        \caption{
            \textbf{Accuracy(\%) of surrogate model} for Llama-70B on text benchmarks.
        }
        \label{tab:surrogate_70b_llm_benchmarks}
    \end{minipage}
    \begin{minipage}[t]{1.\linewidth}
        \captionsetup{type=table}
        \tablestyle{1.18pt}{1.05}
        \begin{tabular}{z{76}|z{20}z{20}z{20}z{20}z{20}z{24}|x{12}}
            encoder fine-tuned on  \ \ \  & \rotatebox{90}{MME$^\text{perc}_\text{binary}$} & \rotatebox{90}{POPE$_\text{binary}$} & \rotatebox{90}{SEED-Bench}
                                          & \rotatebox{90}{MMVet}                           & \rotatebox{90}{LLaVA-Wild}           & \rotatebox{90}{MMB$^\text{en}$} \ \ \
                                          & \rotatebox{90}{performance}                                                                                                    \\
            \shline
            Llama-70B  \ \ \              & 1294                                            & 83.4                                 & 59.8
                                          & 27.0                                            & 45.6                                 & 58.8 \ \ \
                                          &                                                                                                                                \\
            \hline
            surrogate-37B   \ \ \         & 1287                                            & 84.8                                 & 64.2
                                          & 29.6                                            & 54.2                                 & 59.5 \ \ \
                                          &                                                                                                                                \\
            zero-shot grafting \ \ \      & 1315                                            & 86.1                                 & 64.1
                                          & 37.4                                            & 59.7                                 & 60.7 \ \ \
                                          & $\uparrow$                                                                                                                     \\
        \end{tabular}
        \vspace{-1.em}
        \caption{
            \textbf{Accuracy (\%) of encoder fine-tuned by surrogate} for Llama-70B on VLM benchmarks.
        }
        \label{tab:surrogate_70b_vlm_benchmarks}
    \end{minipage}
    \vspace{-1.em}
\end{figure}
%
In Table~\ref{tab:70b_results}, with encoder-only training, our surrogate outperforms the full-size Llama-70B on most VLM benchmarks, except for VisWiz.
This demonstrates the effectiveness of our surrogate models.
The last row shows the performance of zero-shot grafting the surrogate-trained encoder into Llama-70B.
Notably, the performance of zero-shot grafting surpasses the full-size Llama-70B decoder training on some benchmarks by a big margin, demonstrating that our surrogate-trained encoder effectively prompts LLaMA-70B to handle complex visual understanding tasks.
\noindentnewline
Figure~\ref{fig:zero-shot-grafting-llama-70b} presents qualitative results showcasing the strong zero-shot grafting capability of our surrogate-trained encoder, including questions about creativity, negation, and reasoning.
Additionally, Figure~\ref{fig:zero-shot-grafting-llama-70b-ocr} demonstrates its effectiveness on OCR tasks, showing that our surrogate models are able to squeeze robust and detailed visual information into encoders.
%
%
\begin{table*}[t]
    \tablestyle{2.33pt}{1.0}
    \begin{tabular}{z{80}|x{16}x{16}|x{16}x{16}|x{16}x{16}|x{16}x{16}|x{16}x{16}x{16}|x{16}|x{24}|x{20}|x{16}x{16}x{16}|x{18}|x{16}}
                                            & \multicolumn{2}{c}{MME$_\text{binary}$}  & \multicolumn{2}{c}{MME}
                                            & \multicolumn{2}{c}{POPE$_\text{binary}$} & \multicolumn{2}{c}{POPE}
                                            & \multicolumn{3}{c}{SEED-Bench}           & \rotatebox{0}{MM}
                                            & \rotatebox{0}{LLaVA}                     & \rotatebox{0}{MMB}
                                            & \multicolumn{3}{c}{CV-Bench}
                                            & GQA                                      & Vis-                                                     \\
        training method  \ \ \              & cog                                      & perc                     & cog           & perc
                                            & acc.                                     & f1                       & acc.          & f1
                                            & all                                      & img                      & vid
                                            & -Vet                                     & -Wild                    & en
                                            & 2d                                       & 3d                       & avg
                                            &                                          & Wiz                                                      \\
        \shline
        Llama-70B decoder \ \ \             & \textbf{327}                             & \textbf{1545}            & \textbf{345}  & \textbf{1524}
                                            & 84.9                                     & 83.1                     & 84.8          & 82.9
                                            & 63.6                                     & 68.9                     & 43.7
                                            & \textbf{35.5}                            & 67.5                     & \textbf{71.8}
                                            & 61.8                                     & \textbf{73.3}            & \textbf{67.5}
                                            & \textbf{62.4}                            & \textbf{53.0}                                            \\
        \gray{Llama-70B encoder} \ \ \      & \gray{285}                               & \gray{1294}              & \gray{288}    & \gray{1321}
                                            & \gray{83.4}                              & \gray{82.6}              & \gray{82.7}   & \gray{81.2}
                                            & \gray{59.8}                              & \gray{65.4}              & \gray{38.6}
                                            & \gray{27.0}                              & \gray{45.6}              & \gray{58.8}
                                            & \gray{62.2}                              & \gray{59.4}              & \gray{60.8}
                                            & \gray{54.4}                              & \gray{47.4}                                              \\
        \hline
        \gray{surrogate-37B encoder} \ \ \  & \gray{312}                               & \gray{1329}              & \gray{291}    & \gray{1250}
                                            & \gray{85.5}                              & \gray{83.9}              & \gray{86.3}   & \gray{85.0}
                                            & \gray{65.9}                              & \gray{71.1}              & \gray{46.2}
                                            & \gray{28.8}                              & \gray{54.3}              & \gray{63.1}
                                            & \gray{64.7}                              & \gray{64.0}              & \gray{64.3}
                                            & \gray{56.5}                              & \gray{22.7}                                              \\
        zero-shot grafting \ \ \            & 295                                      & 1348                     & 303           & 1298
                                            & \textbf{86.8}                            & \textbf{86.1}            & \textbf{87.0} & \textbf{86.4}
                                            & \textbf{65.4}                            & \textbf{70.7}            & \textbf{45.3}
                                            & 32.8                                     & \textbf{68.9}            & 65.6
                                            & \textbf{63.2}                            & 67.2                     & 65.2
                                            & 51.9                                     & 40.0                                                     \\
    \end{tabular}
    \vspace{-1.em}
    \caption{
        \textbf{Accuracy (\%)} for Llama-70B on VLM benchmarks.
        The \textbf{bold} numbers indicate the best performance between the full-size decoder training and our surrogate-trained encoder by zero-shot grafting.
        A special clarification for LLaVA-Wild is in Sec.~\ref{sec:llava-wild}.
    }
    \label{tab:70b_results}
\end{table*}
%
%
%
\begin{table*}[t]
    \begin{minipage}[t]{\textwidth}
        \tablestyle{2.33pt}{1.0}
        \begin{tabular}{z{32}|z{14}|x{24}|x{16}x{16}|x{16}x{16}|x{16}x{16}|x{16}x{16}|x{16}x{16}x{16}|x{16}|x{24}|x{20}|x{16}x{16}x{16}|x{18}|x{16}}
                                   &                                          & avg.                     & \multicolumn{2}{c}{MME$_\text{binary}$} & \multicolumn{2}{c}{MME}
                                   & \multicolumn{2}{c}{POPE$_\text{binary}$} & \multicolumn{2}{c}{POPE}
                                   & \multicolumn{3}{c}{SEED-Bench}           & \rotatebox{0}{MM}
                                   & \rotatebox{0}{LLaVA}                     & \rotatebox{0}{MMB}
                                   & \multicolumn{3}{c}{CV-Bench}
                                   & GQA                                      & Vis-                                                                                         \\
            method  \ \ \          & X\% \                                    & score
                                   & cog                                      & perc                     & cog                                     & perc
                                   & acc.                                     & f1                       & acc.                                    & f1
                                   & all                                      & img                      & vid
                                   & -Vet                                     & -Wild                    & en
                                   & 2d                                       & 3d                       & avg
                                   &                                          & Wiz                                                                                          \\
            \shline
            baseline   \ \ \       & 10  \ \                                  & 0.5127
                                   & 301                                      & 1178                     & 333                                     & 1139
                                   & 74.5                                     & 73.1                     & 74.3                                    & 73.9
                                   & 46.0                                     & 49.2                     & 34.1                                    & 17.7
                                   & 30.4                                     & 45.6                     & 52.4
                                   & 61.7                                     & 57.1
                                   & 45.4                                     & 43.9                                                                                         \\
                                   & 20 \ \                                   & 0.5153
                                   & 255                                      & 1061                     & 277                                     & 1116
                                   & 71.4                                     & 61.1                     & 71.6                                    & 61.3
                                   & 52.2                                     & 55.5                     & 39.5
                                   & 20.6                                     & 41.3                     & 58.3
                                   & 56.6                                     & 66.8                     & 61.7
                                   & 47.5                                     & 38.5                                                                                         \\
                                   & 30 \ \                                   & 0.6277
                                   & 314                                      & 1447                     & 316                                     & 1399
                                   & 85.5                                     & 84.8                     & 85.5                                    & 84.8
                                   & 59.8                                     & 64.8                     & 40.9
                                   & 30.8                                     & 57.9                     & 66.8
                                   & 60.6                                     & 71.1                     & 65.9
                                   & 57.1                                     & 55.1                                                                                         \\
                                   & 60 \ \                                   & 0.6444
                                   & 353                                      & 1511                     & 358                                     & 1515
                                   & 84.8                                     & 83.1                     & 84.4                                    & 82.4
                                   & 62.4                                     & 67.9                     & 41.5
                                   & 32.4                                     & 64.5                     & 70.9
                                   & 61.5                                     & 72.2                     & 66.8
                                   & 61.3                                     & 48.1                                                                                         \\
                                   & 100 \ \                                  & 0.6538
                                   & 327                                      & 1545                     & 345                                     & 1524
                                   & 84.9                                     & 83.1                     & 84.8                                    & 82.9
                                   & 63.6                                     & 68.9                     & 43.7
                                   & 35.5                                     & 67.5                     & 71.8
                                   & 61.8                                     & 73.3                     & 67.5
                                   & 62.4                                     & 52.9                                                                                         \\
            \hline
            \gray{grafting} \ \ \  & \gray{-} \ \                             & \gray{0.6259}
                                   & \gray{295}                               & \gray{1348}              & \gray{303}                              & \gray{1298}
                                   & \gray{86.8}                              & \gray{86.1}              & \gray{87.0}                             & \gray{86.4}
                                   & \gray{65.4}                              & \gray{70.7}              & \gray{45.3}
                                   & \gray{32.8}                              & \gray{68.9}              & \gray{65.6}
                                   & \gray{63.2}                              & \gray{67.2}              & \gray{65.2}
                                   & \gray{51.9}                              & \gray{40.0}                                                                                  \\
            \hline
            ours   \ \ \           & 10 \ \                                   & 0.6612
                                   & 340                                      & 1404                     & 342                                     & 1430
                                   & 87.3                                     & 86.8                     & 84.9                                    & 82.7
                                   & 67.1                                     & 72.7                     & 45.9
                                   & 37.6                                     & 69.7                     & 70.9
                                   & 66.6                                     & 70.6                     & 68.6
                                   & 60.1                                     & 57.9                                                                                         \\
            \rowcolor{mygray!15}
                                   & 20 \ \                                   & 0.6701
                                   & 369                                      & 1435                     & 361                                     & 1486
                                   & 86.7                                     & 85.6                     & 86.4                                    & 84.9
                                   & 67.2                                     & 72.8                     & 46.0
                                   & 38.8                                     & 70.5                     & 73.2
                                   & 65.7                                     & 74.8                     & 70.3
                                   & 60.6                                     & 52.4                                                                                         \\
                                   & 30 \ \                                   & 0.6704
                                   & 374                                      & 1449                     & 349                                     & 1490
                                   & 87.9                                     & 87.7                     & 86.8                                    & 85.5
                                   & 67.0                                     & 72.0                     & 47.8
                                   & 38.9                                     & 69.3                     & 73.9
                                   & 66.6                                     & 72.8                     & 69.7
                                   & 61.4                                     & 49.2                                                                                         \\
        \end{tabular}
        \vspace{-1.em}
        \caption{
            \textbf{Convergence comparison} with using X\% of training data between baseline and our surrogate training approach for Llama-70B decoder training.
            The \colorbox{mygray!15}{gray row} indicates the training hours reported in the Table \ref{tab:hours} with 20\% of training data for ours.
            See Table \ref{tab:convergence extra} for extra columns with additional benchmarks.
        }
        \label{tab:convergence}
    \end{minipage}
    \vspace{-1.em}
\end{table*}
%
%
\noindentnewline
While the surrogate-trained encoder enables zero-shot conversion of the giant LLM into a VLM, its performance still lags behind that of full-size decoder training.
What benefits can we expect from this surrogate-trained encoder?
Next, we demonstrate that it can accelerate training convergence and improves the performance of full-size decoder training.
\subsection{Reducing Full Decoder Training Cost}
\label{sec:reduce cost}
In the previous sections, we conduct the experiments with a two-stage training strategy, where we simultaneously train the vision adapter in encoder and the translator in decoder during the first stage, and then fine-tune the encoder atop the surrogate in the second stage.
Currently, we are interested in training the full-size decoder, which is the final third training stage.
First, we introduce the training setup, and recipes are introduced in Sec. \ref{sec:training recipes}.
\noindentnewline
\textbf{Models}.
As in previous sections, we use the CLIP-L/14 encoder with an input image size of 336$^2$.
The vision adapter is a two-layer MLP, consisting of consecutive linear layers with a GELU activation in between.
Notably, we maintain a fixed vision adapter size across all model scales, unlike prior works \cite{li2024llava,chen2024internvl} that scale it with model size.
This design choice ensures that variations in adapter size do not introduce unknown effects on the encoder's zero-shot grafting capability, allowing for a controlled initial study.
For state-of-the-art performance, however, the vision adapter can be scaled up with the model size.
The decoders are our surrogate-37B, i.e., $\mathcal{T}(40, 78)$, and full-size Llama-70B.
\noindentnewline
\textbf{Data}.
In the third training stage, the training data is still the same as in the previous two stages -- the LLaVA-1.5-665K \cite{liu2024improved} instructions (without text-only samples).
This choice is based on the following considerations:

\noindent
1) The first training stage focuses on the adapter and translator.
Commonly, vision adapters are trained on captions instead of instructions, but we found no significant difference in experimental outcomes.
Thus, to simplify training, we merge the training of the vision adapter and translator into a single stage using vision-language and text-only instructions.
When forwarding the text-only instructions, gradients backpropagated to the vision adapter are zero.

\noindent
2) The second stage trains encoders with surrogates, aiming to efficiently compress data into encoders while preparing to transfer knowledge to the full-size decoder.
To ensure consistency, we use the same training data for the second and third stages.
It is recommended to use larger and more diverse datasets for those two stages.
\subsubsection{Results: Convergence and Training Cost}
In Table~\ref{tab:convergence}, we compare performance of the typical baseline method and our surrogate training approach across different percentages of training data used for training decoders. The baseline trains Llama-70B with the original CLIP encoder, while ours trains it with our surrogate-trained encoder (the third row in Table~\ref{tab:70b_results}).
First, the gray row represents the performance of zero-shot grafting the surrogate-trained encoder to Llama-70B, which nearly matches the baseline with 30\% of the data.
Second, after training on just 10\% of data, our approach achieves nearly the same performance as the baseline with 100\% of the data, except for MME.
For other benchmarks, our 10\% performance even outperforms the final baseline result.
With continued training, performance remains unchanged, suggesting saturation after 20\% of the data.
In Figure~\ref{fig:cost}, we plot the normalized average score of each X\% data utilization for our method and baseline.
\noindentnewline
We also visualize Table~\ref{tab:hours} in Figure~\ref{fig:cost} for a direct comparison of training hours for each training stage.
First, our pre-training time is longer than the baseline because we train both the vision adapter and the translator with additional text instructions.
However, the key advantage of our surrogate training approach is seen in the decoder training, which is the real bottleneck in common methods.
With 20\% of the data used for our decoder training, we achieve a training cost reduction of $\sim$45\%.
This reduction is the minimum, as our performance in the 10\% case already exceeds the final result of the baseline, as shown in Table~\ref{tab:convergence}.
\begin{table}[h]
    \tablestyle{1.2pt}{1.05}
    \begin{tabular}{z{32}|x{28}|x{28}x{30}x{30}x{36}}
        method  \ \ \   & \# gpu & pre. & ft. enc. & ft. dec. & total hours \\
        \shline
        baseline \ \ \  & 128    & 7.01 & 0.00     & 27.88    & 34.79       \\
        ours   \ \ \    & 128    & 9.36 & 4.25     & \ \ 5.56 & 19.17       \\
    \end{tabular}
    \vspace{-1.em}
    \caption{
        \textbf{Training hours comparison} between baseline and our surrogate approach for training VLMs with Llama-70B, including the time for pretraining (pre.), fine-tuning encoder (ft. enc.), and fine-tuning decoder (ft. dec.).
        Checkpoint loading and saving times are excluded.
        More details in Sec. \ref{sec:training time for llama-70b}.
    }
    \label{tab:hours}
    \vspace{-1.em}
\end{table}
\section{Related Work Overview}
\label{sec:related}
\noindent
\textbf{Understanding LLMs} is a key topic in mechanistic interpretability \cite{sharkey2025open}.
\cite{alain2016understanding} uses linear classifiers (\textit{probes}) to understand the dynamics of intermediate layers in neural networks.
For LLMs, \cite{nostalgebraist} directly employs the output embedding matrix as a probe to classify layer-wise representations, illustrating how input tokens shift from current positions to next ones.
Tuned Lens \cite{belrose2023eliciting} extends this idea with a trainable probe for broader applicability to modern LLMs.
\cite{sun2024transformer} conceptualizes transformer layers as ``painters" that iteratively refine representations and suggests that middle layers share the same representation.
In contrast, we identify two distinct transition phases in LLMs.
\noindentnewline
The shared representation in middle layers suggests redundancy.
\cite{sun2024transformer} further concludes that some middle layers can be removed without a significant performance drop.
\textbf{Prunning LLMs} largely is based on such insight of redundancy.
Notably, both \cite{gromov2024unreasonable} and \cite{men2024shortgpt} found that deep layers are not essential and can be removed.
Interestingly, our surrogate models also replace deep layers in the late phase.
However, our method differs in how we identify the transition point and in our objective.
Unlike prunning, which aims to remove layers while preserving performance, our focus is on the efficiency of surrogate models for encoder transferability.
While our surrogate models consistently underperform compared to their target LLMs, they serve a distinct purpose in producing efficient encoders for VLMs.
\noindentnewline
Our surrogate-trained encoders can directly prompt target LLMs to generate the expected responses without any fine-tuning.
This zero-shot grafting ability aligns with the concept of \textbf{steering LLMs}, a lightweight alternative to fine-tuning LLMs \cite{han2023word, kowsher2024propulsion}.
Prior works show that language models can be guided to perform specific tasks without extensive fine-tuning.
Similarly, in our case, image features from surrogate-trained encoders act as steering tokens, enabling target LLMs to interpret visual content and answer various complicated questions.
\noindentnewline
This capability provides a warm start for further decoder fine-tuning, helping to mitigate the \textbf{expensive training cost of VLMs} \cite{alayrac2022flamingo,li2023blip,wang2023cogvlm,wang2024qwen2,chen2024internvl,tong2025cambrian}.
The costs surged as decoder sizes scale from relatively small models (3B, 8B) to much larger ones, such as 70B \cite{li2024llava}, 110B \cite{li2024llavanext-strong,liu2024llavanext}.
Additionally, increasing the number of image tokens for high-resolution inputs further escalates the computational burden.
LoRA \cite{hu2022lora} could be applied for training VLMs.
While LoRA improves efficiency, it underperforms full fine-tuning, especially in giant LLMs, when applied with small rank (e.g., 8) and alpha (e.g., 32) to query and key decoder layers -- a common practice in LLM training.
Closing this gap needs applying LoRA to {\em entire} transformer layers with large rank and alpha (e.g., rank 128 with alpha 256 as in \cite{li2024llava} for 13B decoder training).
Then LoRA takes about the same time as full decoder fine-tuning.
This limitation likely explains why current VLMs still rely on full decoder fine-tuning.
Critically, contrasting to our surrogate training approach, LoRA does not accelerate convergence.
See more in Sec. \ref{sec:comparing with lora}.
\noindentnewline
Additionally, the idea of using small models to train encoders before applying them to larger decoders has been depicted in \cite{internvl2}.
However, this work is not directly related to ours, as it employs a progressive multi-stage training strategy to just scale up model size and refine image processing from coarse to fine.
No further details are provided on this method, leaving it unclear how it reduces costs.
In contrast, our approach provides a well-defined framework for constructing efficient surrogate models specifically tailored for any target LLM.
Plus, we plug the surrogate-trained encoders directly into their target LLMs, converting them into VLMs without any fine-tuning to perform complex visual understanding tasks.
Further, with our surrogate-trained encoders, the decoder needs only a few full-scale fine-tuning steps to achieve the desired performance.
\vspace{.22em}  
\section{Conclusion}
\label{sec:conclusion}
In this work, we show that vision encoders trained with our surrogate models can accelerate VLM training.
We also note that our surrogate models are not limited to vision encoders.
The main limitation of our approach is the need for a well-designed surrogate, which ideally should be small.
Although our layer-dropping strategy works in principle for any LLM, resulting models are still half the size of their target LLMs, for example, our surrogate-37B for Llama-70B.
This underscores the practical value of surrogate models and highlights the need for ways to create them more efficiently and with better compression.
\clearpage
\setcounter{figure}{0}
\setcounter{table}{0}
\setcounter{equation}{0}
\setcounter{subsection}{0}
\renewcommand{\thesubsection}{A.\arabic{subsection}}
\renewcommand{\thefigure}{A.\arabic{figure}}
\renewcommand{\thetable}{A.\arabic{table}}
\renewcommand{\theequation}{A.\arabic{equation}}
\section*{A. Appendix}
\subsection{Extra Columns for Main Table 8.}
\begin{table}[h]
    \vspace{-.5em}
    \tablestyle{2.33pt}{1.0}
    \begin{tabular}{z{32}|z{14}|x{24}|x{16}|x{16}|x{20}x{20}x{16}|x{16}|x{14}}
                               &                             & avg.               & Text & Doc
                               & \multicolumn{3}{c}{ChartQA} & Info               & AI         \\
        method  \ \ \          & X\% \                       & score
                               & {\scriptsize{VQA}}          & {\scriptsize{VQA}}
                               & overall                     & human              & aug.
                               & {\scriptsize{VQA}}          & {\scriptsize{2D}}               \\
        \shline
        baseline   \ \ \       & 30 \ \                      & 0.2751
                               & 36.9                        & 16.5
                               & 16.9                        & 17.8
                               & 16.1                        & 24.7
                               & 63.6                                                          \\
                               & 60 \ \                      & 0.3041
                               & 46.8                        & 22.2
                               & 17.8                        & 20.2
                               & 15.4                        & 25.9
                               & 64.8                                                          \\
                               & 100 \ \                     & 0.3105
                               & 47.8                        & 22.6
                               & 18.8                        & 21.1
                               & 16.5                        & 25.6
                               & 64.9                                                          \\
        \hline
        \gray{grafting} \ \ \  & \gray{-} \ \                & \gray{0.2821}
                               & \gray{46.2}                 & \gray{21.1}
                               & \gray{16.1}                 & \gray{19.1}
                               & \gray{13.1}                 & \gray{20.7}
                               & \gray{61.2}                                                   \\
        \hline
        ours   \ \ \           & 10 \ \                      & 0.3512
                               & 51.5                        & 29.5
                               & 23.9                        & 25.8
                               & 22.0                        & 28.9
                               & 64.2                                                          \\
        \rowcolor{mygray!15}
                               & 20 \ \                      & 0.3376
                               & 52.3                        & 28.7
                               & 20.3                        & 22.7
                               & 17.8                        & 28.9
                               & 65.7                                                          \\
                               & 30 \ \                      & 0.3468
                               & 53.1                        & 29.1
                               & 21.6                        & 23.7
                               & 19.6                        & 29.4
                               & 66.3                                                          \\
    \end{tabular}
    \vspace{-1.em}
    \caption{
        Extra columns for Table~\ref{tab:convergence} with additional benchmarks.
    }
    \label{tab:convergence extra}
    \vspace{-1.em}
\end{table}
\subsection{Comparing with LoRA}
\label{sec:comparing with lora}
It is reasonable to ask whether applying LoRA \cite{hu2022lora} to full decoder training could reduce training costs enough to eliminate the need for our surrogate training approach.
We evaluate this by applying LoRA to the full Llama-70B decoder training with the same setup as the experiments in Section \ref{sec:reduce cost}.
LoRA is applied to query and key layers in all transformer blocks, with rank 8 and alpha 32, following a common configuration \cite{hu2022lora} for tuning LLMs.
Each training step takes an average of {$14.2$} seconds, including data loading and optimizer steps.
Indeed, LoRA training is faster than full parameter fine-tuning, as shown in Figure~\ref{fig:cost}.
\noindentnewline
However, LoRA cannot reduce training costs as effectively as our surrogate training approach because it does not improve convergence speed.
This happens for two reasons:

\noindent
1) LoRA is designed for fine-tuning an already well-trained model, where the target distribution is mostly aligned.
For example, fine-tuning a {\em language} model on a new {\em text corpus}.
In contrast, VLM training aims to transform a language model from a text-only space to a vision-language space.
Since LoRA updates only a few parameters, it has no sufficient capacity to adapt the language model to vision features, leading to suboptimal performance.

\noindent
2) LoRA reduces trainable model parameters, but not the total training steps.
Conversely, our surrogate training approach speeds up the decoder training convergence by the surrogate-trained encoder, reducing overall training steps.
By addressing optimization efficiency rather than just parameter count, our method is fundamentally more effective.
\subsection{Language Degradation in Decoders}
\label{sec:language degradation}
A common issue in VLM training is that the language decoder underperforms on text benchmarks after training on vision-language instructions.
This degradation is due to the model's focus on vision-language tasks, which can lead to a loss of language understanding.
A typical solution is to mix the text-only corpus with vision-language instructions during training to mitigate this issue.
In our surrogate training approach, we examine whether it can help mitigate this degradation in this ablation.
\begin{table}[t]
    \vspace{.2em}
    \tablestyle{1.18pt}{1.05}
    \begin{tabular}{z{50}|z{18}z{18}z{18}z{18}z{18}z{18}z{18}z{22}|x{14}}
        model \ \ \      & \rotatebox{90}{mmlu}                   & \rotatebox{90}{hellaswag}         & \rotatebox{90}{arc$_\text{easy}$}
                         & \rotatebox{90}{arc$_\text{challenge}$} & \rotatebox{90}{winogrande}        & \rotatebox{90}{piqa}
                         & \rotatebox{90}{boolq}                  & \rotatebox{90}{openbookqa} \ \ \  & \rotatebox{90}{performance}       \\
        \shline
        Llama-70B \ \ \  & 82.6                                   & 86.9                              & 83.4
                         & 71.2                                   & 85.4                              & 83.7
                         & 89.1                                   & 47.6 \ \ \                                                            \\
        \hline
        ours \ \ \       & 82.5                                   & 86.6                              & 84.2
                         & 71.9                                   & 85.3                              & 84.1
                         & 89.7                                   & 47.8 \ \ \                        & -                                 \\
        baseline \ \ \   & 79.4                                   & 84.8                              & 78.9
                         & 69.5                                   & 84.6                              & 83.6
                         & 87.5                                   & 50.2 \ \ \                        & {$\downarrow$}                    \\
    \end{tabular}
    \vspace{-1.em}
    \caption{
        \textbf{Accuracy(\%) of 70B decoders} from the final training stage of our surrogate approach (ours) and baseline method on text benchmarks.
    }
    \label{tab:70b_decoders_llm_benchmarks}
    \vspace{-1.em}
\end{table}
\noindentnewline
In Table~\ref{tab:70b_decoders_llm_benchmarks}, we compare the accuracy of the 70B decoders from the final training stage of our surrogate approach and the baseline method on text benchmarks.
Our decoder retains language performance, matching or even slightly surpassing the original Llama-70B.
In contrast, the baseline method suffers a significant drop in performance on most benchmarks, including MMLU, HellaSwag, ARC, Winogrande, and BoolQ.
\noindentnewline
This is because our surrogate approach fine-tunes the full-size decoder with a few steps on our surrogate-trained encoder, which is already aligned with the LLM's embedding space.
This alignment prevents the decoder's representation from drifting too far, preserving its language understanding.
%
%
\begin{figure}[h]
    \vspace{-1.em}
    \centering
    \includegraphics[width=1.0\linewidth]{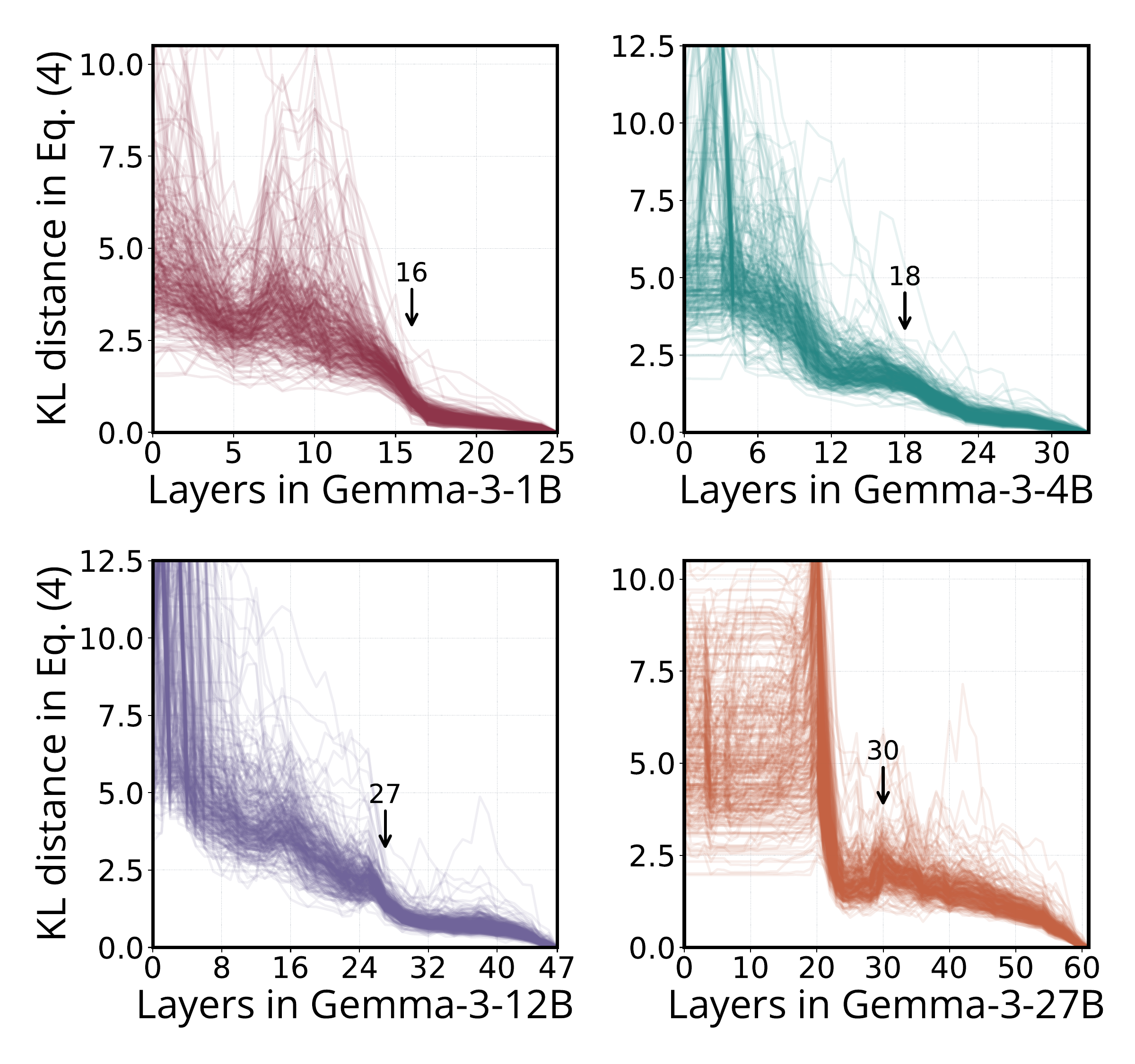}
    \vspace{-1.5em}
    \caption{
        \textbf{The trajectory of prediction} across different layers of 1B, 4B, 12B, and 27B instruct models from Gemma-3.
        The arrow marks the transition point where the trajectories of 300 random samples converge.
    }
    \label{fig:kl-curves-gemma}
    \vspace{-1.em}
\end{figure}
%
%
\begin{table*}[t]
    \tablestyle{2.53pt}{1.0}
    \begin{tabular}{z{40}|x{16}x{16}|x{16}x{16}|x{16}x{16}|x{16}x{16}|x{20}|x{16}|x{24}|x{20}|x{20}|x{18}|x{16}|x{16}|x{16}|x{18}|x{16}|x{16}}
        encoder ft. \ \ \
                  & \multicolumn{2}{c}{MME$_\text{binary}$}  & \multicolumn{2}{c}{MME}
                  & \multicolumn{2}{c}{POPE$_\text{binary}$} & \multicolumn{2}{c}{POPE}
                  & SEED                                     & \rotatebox{0}{MM}
                  & \rotatebox{0}{LLaVA}                     & \rotatebox{0}{MMB}
                  & CV-
                  & GQA                                      & Vis-
                  & Text                                     & Doc
                  & Chart                                    & Info
                  & AI                                                                                \\

        on \ \ \  & cog                                      & perc                     & cog  & perc
                  & acc.                                     & f1                       & acc. & f1
                  & Bench
                  & -Vet                                     & -Wild                    & en
                  & Bench
                  &                                          & Wiz
                  & {\scriptsize{VQA}}                       & {\scriptsize{VQA}}
                  & QA                                       & {\scriptsize{VQA}}
                  & 2D                                                                                \\
        \shline
        $\mathcal{T}(22, 35)$   \ \ \
                  & 236                                      & \ \ 920                  & 284  & 888
                  & 78.0                                     & 80.0                     & 78.0 & 75.1
                  & 43.1                                     & 10.7
                  & 27.9                                     & 20.8
                  & 44.4                                     & 41.6
                  & 12.7                                     & \ \ 7.7
                  & 5.4                                      & \ \ 8.5
                  & \ \ 8.9                                  & 48.1                                   \\
        \hline
        grafting                      \ \ \
                  & 278                                      & 1026                     & 272  & 938
                  & 81.1                                     & 81.6                     & 78.1 & 80.9
                  & 50.1
                  & 20.6                                     & 44.4                     & 49.4
                  & 54.0
                  & 32.9                                     & 44.3
                  & 11.5                                     & 7.6
                  & 10.1                                     & 16.3
                  & 56.5                                                                              \\
    \end{tabular}
    \vspace{-1.em}
    \caption{
        \textbf{Accuracy (\%) of encoder} trained on Qwen3-4B surrogate $\mathcal{T}(22, 35)$ and zero-shot grafted to full-size Qwen3-4B on VLM benchmarks.
        We report only {\mytexttt{all}} for SEED-Bench, {\mytexttt{avg}} for CV-Bench, and {\mytexttt{overall}} for ChartQA to save space.
    }
    \label{tab:qwen3-4b_encoder_acc}
    \vspace{-1.em}
\end{table*}
\subsection{Prediction Trajectory of Gemma-3 Family}
In Figure~\ref{fig:kl-curves-gemma}, we plot the prediction trajectory across different layers of Gemma-1B, 4B, 12B, and 27B instruct models from Gemma-3 \cite{gemma_2025} using 300 random sequences.
Their early phases are instable and spiky, which may be due to the sliding window attention.
But this instability does not affect our method, as noted in Section \ref{sec:prediction trajectory}, where our intial proof-of-concept experiments were conducted on Gemma-2-2B.
Their trajectories still converge at a transition point, marking the shift between early and late phases of the model.
We highlight the transition point for each model.
As the figure shows, the early phase becomes increasingly spiky as model size grows in the Gemma-3 family, while the late phase remains smooth and stable.
\subsection{Prediction Trajectory of Qwen Family}
%
%
\begin{figure}[t]
    \centering
    \includegraphics[width=1.0\linewidth]{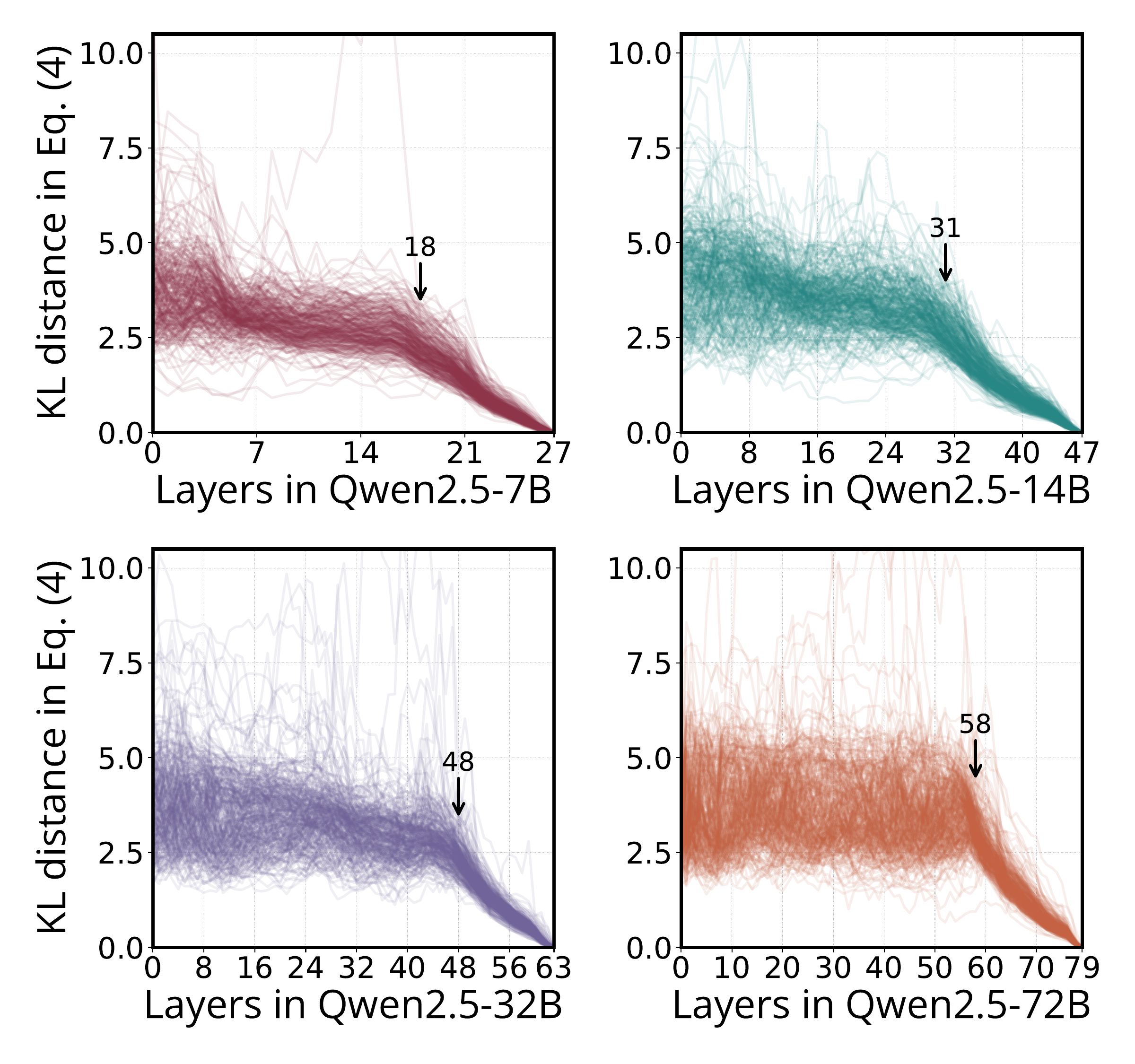}
    \vspace{-1.5em}
    \caption{
        \textbf{The trajectory of prediction} across different layers of 7B, 14B, 32B, and 72B instruct models from Qwen2.5.
        The arrow marks the transition point where the trajectories of 300 random samples converge.
    }
    \label{fig:kl-curves-qwen25}
    \vspace{-1.em}
\end{figure}
%
%
\begin{figure}[t]
    \centering
    \includegraphics[width=1.0\linewidth]{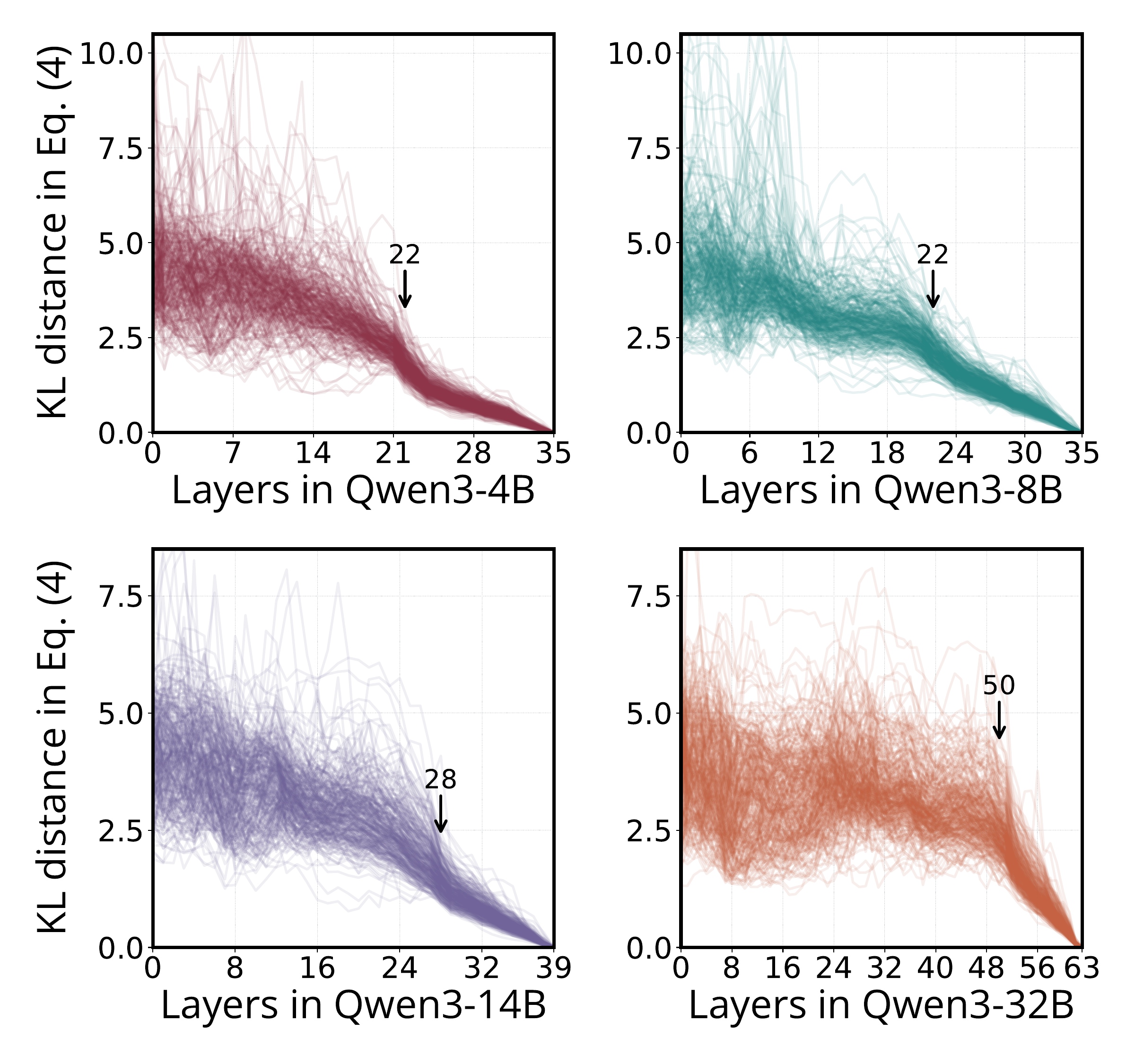}
    \vspace{-1.5em}
    \caption{
        \textbf{The trajectory of prediction} across different layers of 4B, 8B, 14B, and 32B instruct models from Qwen3.
        The arrow marks the transition point where the trajectories of 300 random samples converge.
    }
    \label{fig:kl-curves-qwen3}
    \vspace{-1.em}
\end{figure}
%
In Figure~\ref{fig:kl-curves-qwen25} and Figure~\ref{fig:kl-curves-qwen3}, we plot the prediction trajectory across different layers of Qwen2.5 \cite{yang2025qwen25} and Qwen3 \cite{yang2025qwen3} families.
The difference is that the transition point occurs slightly later in the model, especially for larger models, compared to Llama and Gemma families.
This will cause the surrogate model to be larger than the Llama and Gemma one, for example, the Qwen2.5-72B surrogate model will have 56B parameters.
While this proposed surrogate model is faster and more cost-effective than the full-size 72B model, particularly for extended training, it still remains large for a surrogate we ideally expect.
This limitation has been discussed in Section \ref{sec:conclusion}.
\subsection{Ablation on Method Generalizability}
\label{sec:ablation method generalizability}
To validate the generalizability of our method, we run an ablation on the fresh new Qwen3-4B.
According to the prediction trajectory in Figure~\ref{fig:kl-curves-qwen3}, we build a 2.8B surrogate $\mathcal{T}(22, 35)$ and train the last eight CLIP encoder layers with it under the same experimental settings.
Table \ref{tab:qwen3-4b_encoder_acc} shows the encoder's results on VLM benchmarks:
the first row is with the surrogate;
the second row shows improvements with zero-shot grafting to full-size Qwen3-4B (without thinking mode), except for GQA.
Most improvements are significant.
Thus, our method can generalize well to any pretrained LLM.
\subsection{Ablation on Teacher-Forced Feeding}
\label{sec:ableation teacher-forced}
In Section \ref{sec:prediction trajectory}, the curves in Figure~\ref{fig:kl-curves} are produced using teacher-forced feeding to obtain the model's predictions.
Specifically, we feed 300 randomly sampled pairs of question and response into the model and examine the intermediate feature dynamics.
A potential concern is that this teacher-forced manner with real-world text samples may not accurately represent the model's natural feature dynamics, as it assumes perfect reconstruction of the sequence.
\noindentnewline
To validate our findings, we prompt models with 300 random questions and allow them to predict responses through greedy sampling.
Then we feed these predictions back into the models and find that the resultant curves remain consistent with those obtained from the forced-prediction approach in Figure~\ref{fig:kl-curves}.
\subsection{Training Time for Llama-70B}
\label{sec:training time for llama-70b}
The bottleneck in training Llama-70B is not only the GPU card, but also the network bandwidth for communication.
In our experiments, we use 128 A100-80G GPUs with AWS EFA network.
We shard the 70B parameters across 16 GPU ranks, and the replica group size is 8 using PyTorch FSDP \cite{zhao2023pytorch}.
The batch size is also 128, which means each training step requires communication among all GPUs for forward and backward propagation, without gradient accumulation.
The NCCL communication is AWS EFA\footnote{
    \href{https://aws.amazon.com/hpc/efa/}{\textcolor{urlblue}{Introduction of Amazon Elastic Fabric Adapter (EFA)}}.
}.
For reference of training a full 70B model, the average forward time is {$4.5$} seconds, and the average backward time is {$15.7$} seconds.
Thus, the total average time of each training step is {$\sim$$20.5$} seconds, including the data loading time and optimizer step.
\subsection{Training Recipes}
\label{sec:training recipes}
\begin{figure*}[h]
    \vspace{-.5em}
    \begin{minipage}[t]{.5\textwidth}
        \tablestyle{1.pt}{1.05}
        \begin{tabular}{z{82}|x{64}|x{58}|x{92}}
            recipes \ \                   & stage-1                                                                                           & stage-2                                         & stage-3                       \\
            \shline
            encoder \ \                   & \multicolumn{3}{c}{CLIP-L/14-336px}                                                                                                                                                 \\
            decoder \ \                   & \multicolumn{3}{c}{Llama-3.2-3B, 3.1-8B, 3.1-70B instruct version}                                                                                                                  \\
            adapter in encoder \ \        & \multicolumn{3}{c}{Linear $\rightarrow$ GELU $\rightarrow$ Linear}                                                                                                                  \\
            translator in decoder \ \     & \multicolumn{3}{c}{transformer layer}                                                                                                                                               \\
            \hline
            trainable parameters \ \      & adapter + translator                                                                              & encoder + adapter                               & encoder + adapter + decoder   \\
            learning rate \ \             & 1e-4                                                                                              & 5e-5                                            & 2e-5 for 70B, 5e-5 for others \\
            batch size \ \                & 256                                                                                               & \multicolumn{2}{c}{128}                                                         \\
            instruction datasets \ \      & \ GenQA\cite{chen2024genqa}-500K                                                                  & \multicolumn{2}{c}{LLaVA-1.5-665K}                                              \\
                                          & LLaVA-1.5-665K                                                                                    & \multicolumn{2}{c}{(ShareGPT-40K not included)}                                 \\
            translator layer index \ \    & \multicolumn{3}{c}{$16$ for Llama-3B, $17$ for Llama-8B, and $40$ for Llama-70B}                                                                                                    \\
            image input size \ \          & \multicolumn{3}{c}{fixed size of 336$^2$ pixels}                                                                                                                                    \\
            image augmentation \ \        & \multicolumn{3}{c}{pad to input size with per-channel mean pixel value}                                                                                                             \\
            number of A100-80G \ \        & \multicolumn{3}{c}{16 (Llama-3B, -8B) and 128 (Llama-70B)}                                                                                                                          \\
            warmup ratio \ \              & \multicolumn{3}{c}{3\% of total batch iterations}                                                                                                                                   \\
            lr scheduler \ \              & \multicolumn{3}{c}{cosine annealing with $\text{lr}\_{\text{min}} = 0$}                                                                                                             \\
            optimizer \ \                 & \multicolumn{3}{c}{AdamW($\beta_1 = \text{0.9}, \beta_2 = \text{0.999}, \text{eps}=\text{1e-8}$)}                                                                                   \\
            weight decay \ \              & \multicolumn{3}{c}{0}                                                                                                                                                               \\
            gradient clip \ \             & \multicolumn{3}{c}{max$\_$norm = 1.0 with $2$-norm}                                                                                                                                 \\
            epochs \ \                    & \multicolumn{3}{c}{1}                                                                                                                                                               \\
            \hline
            precision \ \                 & \multicolumn{3}{c}{ bfloat16}                                                                                                                                                       \\
            PyTorch FSDP \ \              & \multicolumn{3}{c}{enabled}                                                                                                                                                         \\
            gradient accumulation \ \     & \multicolumn{3}{c}{enabled}                                                                                                                                                         \\
            activation checkpointing \ \  & \multicolumn{3}{c}{enabled}                                                                                                                                                         \\
        \end{tabular}
        \vspace{-1.em}
        \captionsetup{type=table}
        \caption{
            \textbf{Training hyper-parameters, recipes and settings}.
        }
        \label{tab:hyper params}
    \end{minipage}
    \hspace{6.0em}
    \begin{minipage}[h]{.37\textwidth}
        \vspace{7.em}
        \includegraphics[width=1.\linewidth]{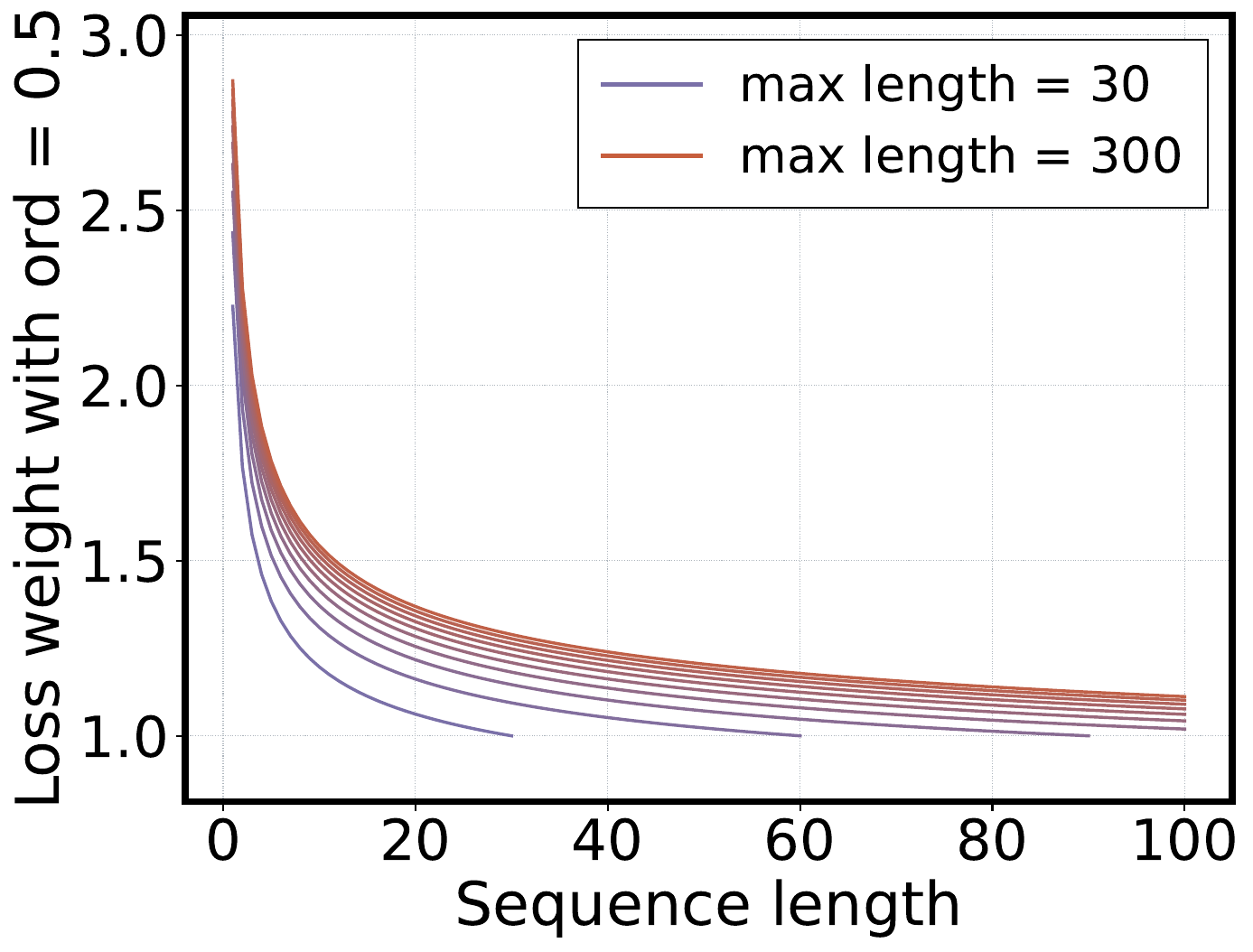}
        \vspace{-1.5em}
        \caption{
            \textbf{Dynamic loss weights} for balancing loss contributions from different response lengths in a global batch.
            Multiple curves represent different maximum response lengths in a batch, gradually increasing from 30 to 300.
        }
        \label{fig:balance loss weights}
    \end{minipage}
    \vspace{-1.em}
\end{figure*}

In the analysis experiments of Section~\ref{sec:method}, we train the entire encoder with surrogate models for one epoch during the second training stage.
In contrast, the main experiments in Section~\ref{sec:generalizing} follow a different setting: only the last eight layers of the encoder are trained for one epoch, using either surrogate-37B or full-size Llama-70B, while the remaining layers are frozen.
This setting is also applied to the baseline method.
Aside from this difference, all other training recipes remain the same.
Next, we provide a detailed description of the training recipes.
\noindentnewline
We apply the chat template specific to each LLM, including any special chat tokens, to the input conversations.
For instance, the dialog shown in Figure~\ref{fig:zero-shot-grafting-llama-8b} starts as raw text; after applying the chat template, it becomes:
%
\begin{figure}[h]
    \centering
    \vspace{-.5em}
    \includegraphics[width=1.\linewidth]{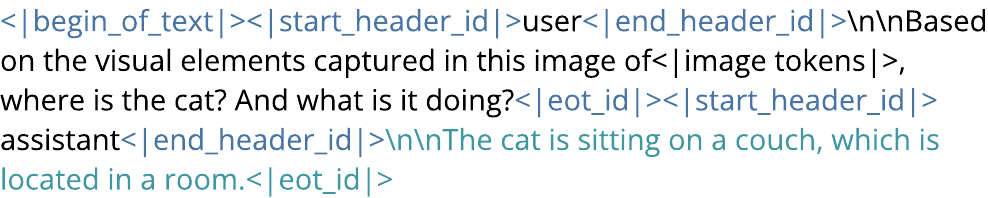}
    \label{fig:input format}
    \vspace{-2.5em}
\end{figure}
%
\noindentnewline
In VLM training, cross-entropy loss for next-token prediction is typically applied only to the green tokens in responses.
The special token {\mytexttt{eot\_id}} marks the end of a conversation turn, while all other tokens are masked out.
However, we found that during encoder-only training, the loss should also be applied to the blue special tokens.
Without this adjustment, the encoder struggles to properly follow question instructions and generate desired responses, both in zero-shot grafting senarios and when paired with surrogate models.
For Llama-70B, along with their surrogates, we fully fine-tune all parameters during decoder training.
Hyper-parameters are in Table~\ref{tab:hyper params}.
%
%
%
%
\subsection{Evaluation Benchmarks}
We evaluate VLMs following LLaVA-1.5~\cite{liu2024improved} with additional benchmarks, including
MME~\cite{fu2023mme},
POPE~\cite{li2023evaluating},
LLaVA-in-the-Wild~\cite{liu2024visual},
SEED-Bench~\cite{li2023seed},
MM-Vet~\cite{yu2023mm},
MMBench~\cite{liu2025mmbench},
TextVQA~\cite{singh2019towards},
GQA~\cite{hudson2019gqa},
DocVQA~\cite{mathew2021docvqa},
ChartQA~\cite{masry2022chartqa},
InfoVQA~\cite{mathew2022infographicvqa},
AI2D~\cite{kembhavi2016diagram},
Viz-Wiz~\cite{gurari2018vizwiz}
using {\mytexttt{lmms-eval}} toolkit~\cite{lmms-eval2024}.
We also evaluate models on the vision-centric benchmark CV-Bench \cite{tong2025cambrian}.
For language models, we evaluate them on
MMLU~\cite{hendrycks2020measuring},
HellaSwag~\cite{zellers2019hellaswag},
ARC~\cite{clark2018think},
PIQA~\cite{bisk2020piqa},
Winogrande~\cite{sakaguchi2021winogrande},
BoolQ~\cite{clark2019boolq},
and OpenBookQA~\cite{mihaylov2018can}
using {\mytexttt{lm-harness}} toolkit~\cite{eval-harness}.
The few-shot setting and the type of reported accuracy for text benchmarks in {\mytexttt{lm-harness}} is shown in Table~\ref{tab:few-shot setting on text benchmarks}.
\begin{table}[h]
    \tablestyle{1.18pt}{1.05}
    \begin{tabular}{z{55}|x{19.5}x{19.5}x{19.5}x{19.5}x{19.5}x{19.5}x{19.5}x{19.5}}
        \ \ \                  & \rotatebox{90}{mmlu}                   & \rotatebox{90}{hellaswag}  & \rotatebox{90}{arc$_\text{easy}$}
                               & \rotatebox{90}{arc$_\text{challenge}$} & \rotatebox{90}{winogrande} & \rotatebox{90}{piqa}
                               & \rotatebox{90}{boolq}                  & \rotatebox{90}{openbookqa}                                     \\
        \shline
        number of shots \ \ \  & 5                                      & 10                         & 0 \ \
                               & 25                                     & 5 \ \                      & 0 \ \
                               & 0 \ \                                  & 0 \ \                                                          \\
        acc. type \ \ \        & -                                      & norm                       & norm
                               & norm                                   & -                          & norm
                               & -                                      & norm                                                           \\
    \end{tabular}
    \vspace{-1.em}
    \caption{
        \textbf{Few-shot setting} for text benchmarks in {{\mytextttforfootnote{lm-harness}}}.
        For accuracy type, ``norm'' refers to length-normalized accuracy.
    }
    \label{tab:few-shot setting on text benchmarks}
    \vspace{-1.5em}
\end{table}
\subsection{Surrogate Training for Smaller Models}
For interested readers, we share our experience applying our surrogate training approach to smaller language models, such as Llama-3B and 8B, compared with Llama-70B.
Before diving in, our key takeaway is: \textit{our surrogate training approach is most effective for giant LLMs.}
The larger the LLM decoder and the training data scale in VLMs, the greater the cost reduction our method achieves.
\noindentnewline
Applying this approach to relatively small LLMs is unnecessary, two reasons:

\noindent
a) Their training costs are already affordable nowadays.

\noindent
b) It introduces additional hyper-parameters with minimal cost savings.
\noindentnewline
This section serves purely as a discussion, as our experiments revealed some interesting yet unverified observations.
We share these findings to provide insight into potential limitations and edge cases of our method, which may inform future research.
%
%
\begin{figure}[h]
    \vspace{.5em}
    \centering
    \includegraphics[width=1.\linewidth]{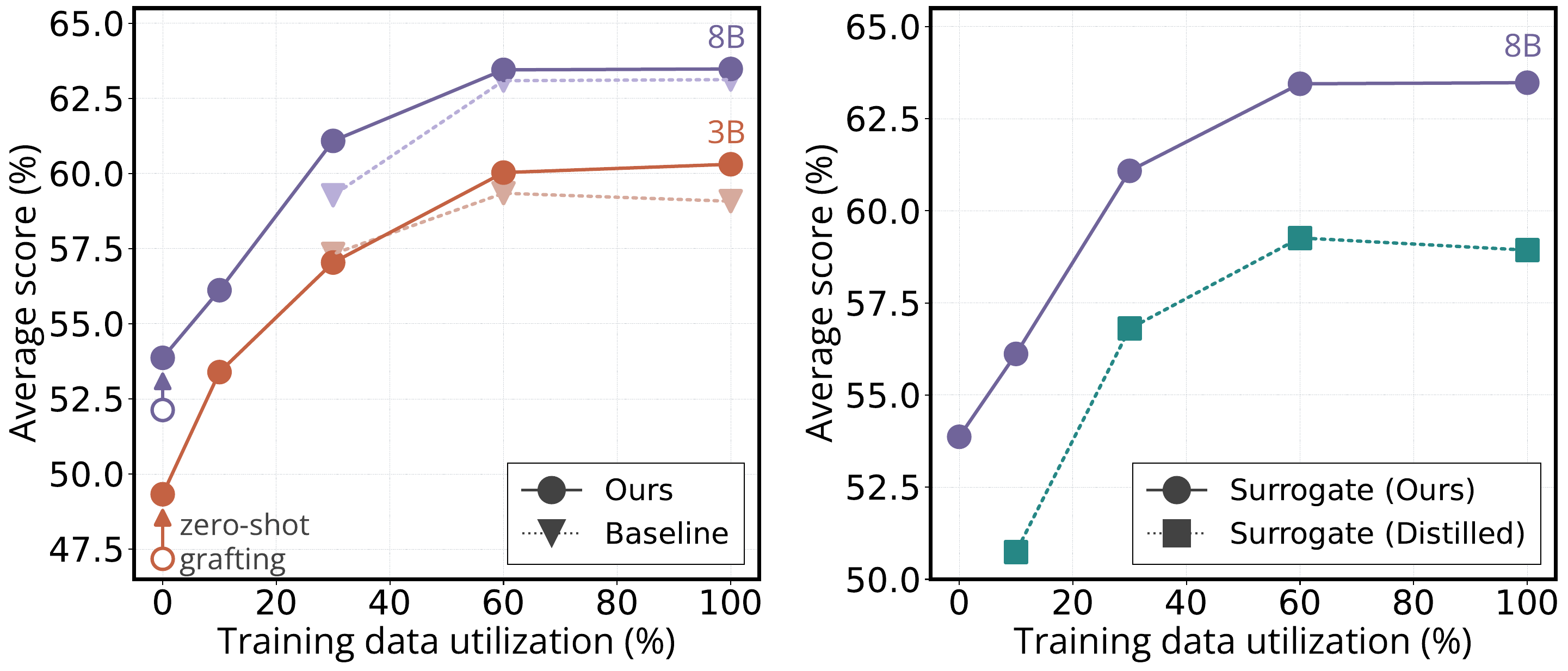}
    \vspace{-1.5em}
    \caption{
        \textbf{Training cost comparison} among our surrogate training approach, baseline method, and training with distill models as surrogates for Llama-3B and 8B.
    }
    \label{fig:cost small}
    \vspace{-1.5em}
\end{figure}
%
%
\noindentnewline
When applying our surrogate training approach to Llama-3B and 8B, we observe performance degradation in the third training stage, particularly on benchmarks requiring a short answer (e.g., ``yes'' or ``no'' in MME and POPE) or single word/phrase (e.g., GQA and VizWiz).
The initial thought is that we may encounter the overfitting issue since we use the same LLaVA-1.5-665K instructions for all the training stages.
\noindentnewline
After further investigation, we find that the performance degradation is not due to overfitting.
Instead, our surrogate is highly effective at training encoders with strong zero-shot grafting capability in the second training stage.
Then in the third stage, the encoder triggers the target LLM to generate responses that already align with the training data distribution, resulting in lower-than-expected loss values, especially for short answers or single words/phrases.
This issue arises from how loss is mean-reduced in a batch -- by dividing the total loss by the number of tokens involved (e.g., green tokens in responses).
As a result, the loss from short answers is overwhelmed by that of longer answers, such as open-ended responses with hundreds of tokens, leading to insuffcient gradient updates for short answers.
\noindentnewline
To address this issue and verify our hypothesis, we adjust the loss calculation to consider the number of tokens in responses.
Specifically, we multiply the loss for each response group by a \textit{dynamic} loss weight to balance the loss contributions from different response lengths within a batch.
Asume the length of the $i$-th response is $L_i$ and the total number of responses in a \textit{global} batch is $N$ across all ranks.
The dynamic loss weight for the $i$-th response is calculated as:
\begin{align*}
    w_i = \left( \frac{\max_{j \in \{1, ..., N\}} L_j}{\text{log} \ L_i}\right)^\text{ord},
    w_i \leftarrow \frac{w_i}{(\sum_{j=1}^{N} w_j) / (\sum_{j=1}^{N} L_j)}.
\end{align*}
We then scale the loss for all tokens in the $i$-th response by $w_i$.
Figure~\ref{fig:balance loss weights} illustrates how this weighting adjusts based on the maximum response length in a batch.
In our experiments, we set $\text{ord} = 0.5$.
With this adjustment, we solve the performance degradation issue and achieve comparable performance to the baseline method, as shown in the left subfigure of Figure~\ref{fig:cost small}.
We believe dynamic loss weighting is not the only solution.
The issue could be addressed by using {\em more diverse, larger-scale datasets} in the second and third stages, unlike the single small set of vision-language instructions used in our experiments.
\begin{table*}[t]
    \centering
    \begin{minipage}[t]{1.\textwidth}
        \tablestyle{1.2pt}{1.15}
        \begin{tabular}{z{135}|y{110}|y{243}}
            input question \ \ \                             & \ \ \ response of model $\mathcal{T}(1, 16)$                  & \ \ \ response of model $\mathcal{T}(16, 24)$                                                             \\
            \shline
            Which one is larger, the cat or the dog? \ \ \   & \ \ \ The cat is larger than the dog.                         & \cellcolor{mygray!15} \ \ \ The dog is larger.                                                            \\
            1 is larger than 10, is it correct? \ \ \        & \ \ \ No, 10 is not larger than 1.                            & \cellcolor{mygray!15} \ \ \ The statement is not true. The number 1 is less than the number 10.           \\
            Happy birthday to you! \ \ \                     & \ \ \ Happy birthday to you!                                  & \cellcolor{mygray!15} \ \ \ Thank you for the birthday greetings! I am happy to celebrate with you. \ \ \ \\
            I love you. \ \ \                                & \ \ \ I'm glad to hear that. I'm here                         & \cellcolor{mygray!15} \ \ \ I am an AI language model and do not have personal feelings or                \\
                                                             & \ \ \ to help you with any questions                          & \cellcolor{mygray!15} \ \ \ emotions. However, I can provide you with information and assistance in       \\
                                                             & \ \ \ or tasks you may have.                                  & \cellcolor{mygray!15} \ \ \ your own way.                                                                 \\
            1, 2, 3, 4. Count the next three numbers. \ \ \  & \ \ \ 1, 2, 3, 4                                              & \cellcolor{mygray!15} \ \ \ 1, 2, 3, 4, 5, 6, 7, 8, 9, 10                                                 \\
            \hline
            What is 100 + 13 = ? \ \ \                       & \cellcolor{mygray!15} \ \ \ 100 + 13 = 113                    & \ \ \ 100 + 13 = 125                                                                                      \\
            Calculate 2 * 10.2 = ? \ \ \                     & \cellcolor{mygray!15} \ \ \ 2 * 10.2 = 20.4                   & \ \ \ 2 * 10.2 = 22.4                                                                                     \\
            Calculate 1 + 10 = ? \ \ \                       & \cellcolor{mygray!15} \ \ \ 1 + 10 = 11                       & \ \ \ 1 + 10 = 20                                                                                         \\
            Is 8.11 {$>$} 8.3? \ \ \                         & \cellcolor{mygray!15} \ \ \ No, 8.11 is not greater than 8.3. & \ \ \ Yes, 8.11 {$>$} 8.3.                                                                                \\
        \end{tabular}
        \vspace{-1.em}
        \caption{
            \textbf{Simple ability assessment} for two small models of Gemma-2B.
            The model $\mathcal{T}(1, 16)$ means the trained translator replaces the early transition phase, while $\mathcal{T}(16, 24)$ replaces the late phase.
            Response is sampled with greedy decoding.
        }
        \label{tab:reasoning ability}
    \end{minipage}
    \vspace{-1.em}
\end{table*}
\subsection{Distilled Model as Surrogate}
In some LLM families, such as Llama \cite{dubey2024llama} and Gemma \cite{gemma_2024}, they distill small models (Llama-1B) with large ones (Llama-8B).
We investigate whether these distilled models can serve as surrogates for training encoders.
The core issue with using distilled models as surrogates is that they have different embedding dimensions from the larger models.
So additional training stage is required to align the embedding dimensions, e.g., with a linear layer.
We conduct the experiments on Llama-1B and 8B, treating Llama-1B as a surrogate model.
The extra training stage is added between the second and third stages, where we fix the surrogate-trained encoder and other parameters, only training the linear layer to align the embedding dimensions.
In the right subfigure of Figure~\ref{fig:cost small}, we compare the performance of the distilled model as a surrogate with our surrogate training approach using the same training setup, including the $\text{ord} = 0.5$ dynamic loss weighting.
The distilled model as a surrogate performs worse than our surrogate training approach, indicating that distilled models are not ideal surrogate choices for our method.
\subsection{Performance Drop on LLaVA-Wild}
\label{sec:llava-wild}
Our models and the baseline model both score lower on the LLaVA-in-the-Wild \cite{liu2024visual} than the official scores reported in LLaVA-1.5 \cite{liu2024improved}, appearing in Tables \ref{tab:70b_results} and \ref{tab:convergence}, as well as in the ablation studies in Section \ref{sec:method}.
This drop is expected due to a change in the GPT-4 API model version in the juding process.
LLaVA-1.5 uses {\mytexttt{GPT4-0314}}, which has been deprecated.
Instead, the evaluation toolkit {\mytexttt{lmms-eval}} uses {\mytexttt{GPT4-0613}}, which systematically results in lower scores across all models.
For example, LLaVA-1.5-7B drops from $65.3$ to $59.6$, and LLaVA-1.5-70B from $72.8$ to $66.1$.
For more details, please refer to README.md\footnote{
    \href{https://docs.google.com/spreadsheets/d/1a5ImfdKATDI8T7Cwh6eH-bEsnQFzanFraFUgcS9KHWc/edit?gid=0\#gid=0}{\textcolor{urlblue}{{\mytextttforfootnote{lmms-eval}} - Comprehensive evaluation results of LLaVA family models.}}
} in {\mytexttt{lmms-eval}} repository.
\subsection{Different Abilities in Two Transition Phases}
\label{sec:ability emerges}
In our initial experiments with Gemma-2-2B for Section~\ref{sec:method}, we observe that different abilities emerge in the two transition phases of the model.
\noindentnewline
Gemma-2B has 26 layers and its transition point is at layer 16, as shown in Figure~\ref{fig:kl-curves}.
We construct two small models by replacing transition phases with a translator: $\mathcal{T}(1, 16)$ to replace the early phase, and $\mathcal{T}(16, 24)$ to replace the late phase.
As in the ablation studies of Section~\ref{sec:method}, we train the translator for one epoch in the first training stage.
When we prompt these two small models with the same question, we find that $\mathcal{T}(16, 24)$ and $\mathcal{T}(1, 16)$ exhibit different abilities in their responses.
\noindentnewline
In Table~\ref{tab:reasoning ability}, we show the responses of both models to a set of questions.
The first block includes simple questions that test basic common sense reasoning, factual recall, and conversational coherence.
Model $\mathcal{T}(16, 24)$ performs well on these questions, in which the early phase is preserved.
It can correctly answers that the dog is larger than the cat, provides a coherent response to the birthday greeting, and appropriately declines the love confession as an AI model.
Additionally, it follows the instruction to count the next numbers but misinterprets how many
to include.
In contrast, model $\mathcal{T}(1, 16)$ struggles with these questions, in which the late phase is preserved.
This suggests that:

\noindent
a) Common sense reasoning, factual knowledge, and conversational abilities are primarily stored in the early-phase parameters.

\noindent
b) Despite training the translator in $\mathcal{T}(1, 16)$, it may not fully recover the model's knowledge, possibly due to its limited capacity with fewer parameters than the full model.
\noindentnewline
The second block of questions tests the model's ability to perform basic arithmetic calculations and comparisons.
Conversely, $\mathcal{T}(1, 16)$ correctly handles even floating-point multiplication, while $\mathcal{T}(16, 24)$ fails entirely.
This suggests that arithmetic computation and numerical comparison are primarily handled by the late-phase parameters.
\vspace{.6em}  
\subsection{Potential Use Cases}
Our surrogate training approach may also benefit large encoder training, instead of the full-size decoder, in the third stage.
For example, InternViT \cite{chen2024internvl} trains large encoders (6B) with LLMs.
Using a surrogate to align the encoder beforehand could provide a warm start for continued training with the frozen full-size LLM, reducing overall cost.
This is feasible since the loss function remains consistent (with a contrastive term), and the encoder is strongly aligned with the full-size LLM via the surrogate like in our experiments.
\clearpage
\section*{Acknowledgements}
We sincerely appreciate Meta's support in providing GPUs for our experiments.
The authors at University of Maryland were supported by DAPRA TIAMAT, the ONR MURI program, the National Science Foundation (IIS-2212182), and the NSF TRAILS Institute (2229885).
Additional commercial support was provided by the Amazon Research Award program, Open Philanthropy, and Capital One Bank.
\section*{Disclaimer}
The image-based training data was used only to train vision encoders to produce image features, not generative components.
    {
        \small
        \bibliographystyle{ieeenat_fullname}
        \bibliography{main}
    }

\end{document}